\documentclass[sigconf]{acmart}

\usepackage{amsmath}    
\usepackage{algorithm}
\usepackage{algorithmic}
\usepackage{enumitem}
\usepackage{multirow}
\usepackage{booktabs}
\usepackage{array}
\usepackage{subcaption} 
\usepackage{makecell}
\usepackage{xcolor}
\AtBeginDocument{%
  }

\setcopyright{acmlicensed}
\copyrightyear{2018}
\acmYear{2018}
\acmDOI{XXXXXXX.XXXXXXX}
\acmConference[Conference acronym 'XX]{Make sure to enter the correct
  conference title from your rights confirmation email}{June 03--05,
  2018}{Woodstock, NY}
\acmISBN{978-1-4503-XXXX-X/2018/06}

\begin{document}

\title{HeteroHBA: A Generative Structure-Manipulating Backdoor Attack on Heterogeneous Graphs}


\author{Honglin Gao}
\affiliation{%
  \department{School of Electrical and Electronic Engineering}
  \institution{Nanyang Technological University}
  \country{Singapore}}
\email{honglin001@e.ntu.edu.sg}

\author{Lan Zhao}
\affiliation{%
  \department{School of Electrical and Electronic Engineering}
  \institution{Nanyang Technological University}
  \country{Singapore}}
\email{zhao0468@e.ntu.edu.sg}

\author{Ren Junhao}
\affiliation{%
  \department{School of Electrical and Electronic Engineering}
  \institution{Nanyang Technological University}
  \country{Singapore}}
\email{JUNHAO002@e.ntu.edu.sg}

\author{Xiang Li}
\affiliation{%
  \department{School of Electrical and Electronic Engineering}
  \institution{Nanyang Technological University}
  \country{Singapore}}
\email{xiang002@e.ntu.edu.sg}

\author{Gaoxi Xiao}
\affiliation{%
  \department{School of Electrical and Electronic Engineering}
  \institution{Nanyang Technological University}
  \country{Singapore}}
\email{egxxiao@ntu.edu.sg}






\renewcommand{\shortauthors}{Trovato et al.}

\begin{abstract}
Heterogeneous graph neural networks (HGNNs) have achieved strong performance in many real-world applications, yet targeted backdoor poisoning on heterogeneous graphs remains less studied. We consider backdoor attacks for heterogeneous node classification, where an adversary injects a small set of trigger nodes and connections during training to force specific victim nodes to be misclassified into an attacker-chosen label at test time while preserving clean performance. We propose HeteroHBA, a generative backdoor framework that selects influential auxiliary neighbors for trigger attachment via saliency-based screening and synthesizes diverse trigger features and connection patterns to better match the local heterogeneous context. To improve stealthiness, we combine Adaptive Instance Normalization (AdaIN) with a Maximum Mean Discrepancy (MMD) loss to align the trigger feature distribution with benign statistics, thereby reducing detectability, and we optimize the attack with a bilevel objective that jointly promotes attack success and maintains clean accuracy. Experiments on multiple real-world heterogeneous graphs with representative HGNN architectures show that HeteroHBA consistently achieves higher attack success than prior backdoor baselines with comparable or smaller impact on clean accuracy; moreover, the attack remains effective under our heterogeneity-aware structural defense, CSD. These results highlight practical backdoor risks in heterogeneous graph learning and motivate the development of stronger defenses.
\end{abstract}


\received{20 February 2007}
\received[revised]{12 March 2009}
\received[accepted]{5 June 2009}

\maketitle

\section{Introduction}
Graph-structured data pervade diverse application domains, ranging from social networks \citep{gnnsocialnet,GNNSocialNet2} and signal processing \citep{gnnsingalprocess} to biological networks \citep{gnnbiological} and knowledge graphs \citep{gnnknowledgegraph}. 
Unlike homogeneous graphs, heterogeneous graphs (HGs) integrate multiple node and edge types to model complex real-world relationships, such as the intricate web of researchers, papers, and institutions in academic networks. 
This structural flexibility enables HGs to serve as the backbone for critical systems, including recommendation engines and financial risk modeling \citep{recommandsystem, financeriskassesment}. 
Heterogeneous Graph Neural Networks (HGNNs) \citep{HGNNNC1, HGNNNC2, HGNNLP1, HGNNLP2} have emerged as the dominant paradigm for leveraging this diverse relational information, demonstrating exceptional efficacy in tasks such as fraud detection \citep{heteorgeneousFinance1, heteorgeneousFinance2} and personalized cross-domain recommendation \citep{heterogeneousRecommendationSystem1, heterogeneousRecommendationSystem2}.

Despite these functional advancements, the predominant research focus on performance maximization has left the security landscape of HGNNs relatively underexplored. 
Recent studies highlight a concerning susceptibility to adversarial threats \citep{heterogeneousGraphAttack1, heterogeneousGraphAttack2}, among which backdoor attacks pose a particularly severe risk due to their inherent stealthiness and potential to subvert critical decision-making \citep{GraphBackdoor}. 
While the study of backdoor mechanisms involving the manipulation of models via intentionally altered training data has been exhaustive in computer vision \cite{CVBackdoor1, CVBackdoor2} and natural language processing \cite{nlpBackdoor1,nlpBackdoor2}, investigation within the HGNN domain remains nascent. 
The complex structure of HGNNs, while endowing them with powerful capabilities, simultaneously introduces new security vulnerabilities. The rich semantic interactions between distinct node types expose a broader attack surface, enabling attackers to implant backdoors through the precise tampering of structures or features. 
Once implanted, such hidden triggers can disastrously mislead the model. For instance, in a financial ecosystem, an attacker might inject a fraudulent account with subtle connections to transaction histories, manipulating the system's judgment to evade detection and cause severe security breaches.

However, despite these imminent threats, the exploration of backdoor attacks in the heterogeneous domain remains limited. 
Existing literature has predominantly focused on homogeneous graphs, where node and edge types are uniform. 
For instance, Dai et al. \cite{UGBA} and Zhang et al. \cite{DPGBA} identify that naive trigger insertion often disrupts the message-passing process, compromising attack stealthiness. 
To mitigate this, they propose the Unnoticeable Graph Backdoor Attack (UGBA) and Distribution Preserving Graph Backdoor Attack (DPGBA), which employ a bi-level optimization framework to execute attacks within a constrained budget while minimizing detectability. 
Alternatively, Xing et al. \cite{CGBA} introduce the Clean-Label Graph Backdoor Attack (CGBA), a strategy that injects triggers into feature space without altering labels or graph structure. 
By selecting triggers highly similar to neighbor features, CGBA preserves structural integrity and resists defense mechanisms. 
Nevertheless, these homogeneous-centric approaches fundamentally overlook the rich semantics of edge relationships and node types inherent in heterogeneous graphs. 
Directly applying them to HGNNs fails to exploit the distinct vulnerabilities arising from heterogeneity. 
While a recent study, HGBA \cite{HBA}, attempts to target heterogeneous graphs, it relies on static triggers. This lack of diversity leads to rigid patterns that fail to mimic benign patterns.
To bridge this gap, we propose a novel \underline{Hetero}geneous \underline{H}ierarchical \underline{B}ackdoor \underline{A}ttack (HeteroHBA), which adaptively generates diverse backdoors for poisoning.

To address the limitations of existing baselines, we propose HeteroHBA, a targeted backdoor attack framework for heterogeneous graphs that generates stealthy, instance-adaptive triggers. HeteroHBA first selects influential auxiliary neighbors via an explanation-based screening mechanism, then leverages our proposed GraphTrojanNet to generate trigger nodes with diverse connections. For statistical stealthiness, AdaIN is used during trigger generation to align feature statistics, and after the bi-level optimization converges, we further apply a lightweight IDA-AT refinement module guided by an MMD loss to better match the benign feature distribution.

Our contributions are threefold, summarized as follows.
\begin{itemize}
    \item Novel perspective on heterogeneous backdoors: We study targeted backdoor vulnerabilities of HGNNs in heterogeneous settings and propose HeteroHBA, which goes beyond static triggers by leveraging structural saliency and type-aware context to generate instance-adaptive, stealthy triggers.
    \item Adaptive generative mechanism: We introduce GraphTrojanNet to generate victim-aware trigger features and adaptive connection patterns, and combine random masking, a diversity regularizer, and bi-level optimization to mitigate structural collapse and improve efficacy across heterogeneous schemas.
    \item Statistical stealthiness and refinement: We apply AdaIN during generation to align trigger feature statistics with benign distributions, and further refine triggers with a post-optimization module, IDA-AT, guided by an MMD-based objective to enhance distribution-level stealthiness.
    \item Comprehensive evaluation and defense baseline: Experiments on three benchmark datasets show that HeteroHBA achieves consistently strong ASR while preserving clean performance, and we propose a Cluster-based Structural Defense (CSD) as a heterogeneous defense baseline for robustness evaluation.
\end{itemize}

The remainder of this paper is structured as follows: Section 2 provides a brief review of related work. Section 3 introduces the necessary preliminaries and definitions. Our proposed methodology is presented in detail in Section 4. To evaluate its effectiveness, we conduct extensive experiments and analyses on multiple benchmark datasets and models in Section 5. Finally, Section 6 concludes the paper.

\section{Related Work}

\subsection{Heterogeneous Graph Neural Networks}
HGNNs have evolved significantly in recent years \citep{RHGNN,HGCN,MAGNN,RpHGNN}, with various architectures designed to effectively capture heterogeneous relationships. Among them, the most representative models include a few as follows. HAN \citep{HAN} introduces meta-path-based attention to selectively aggregate information along predefined relational paths, providing interpretability in node representations. HGT \citep{HGT} extends this approach by leveraging a transformer-based architecture to dynamically model heterogeneous interactions. Meanwhile, SimpleHGN \citep{SimpleHGN} optimizes message passing by simplifying the heterogeneity modeling process, making it computationally efficient while maintaining strong performance. 

While these methods significantly improve learning on heterogeneous graphs, their robustness under malicious manipulation has received limited attention. The unique characteristics of HGNNs, such as diverse node and edge types and advanced attention mechanisms, present both opportunities and challenges for potential attackers, making them an important area for further exploration.

\subsection{Backdoor Attack on Homogeneous Graph}
Backdoor attacks on graph neural networks (GNNs) embed hidden triggers during training, allowing adversaries to control outputs under specific conditions. Existing attacks on homogeneous graphs are classified by their trigger injection strategies.

\textbf{Feature-based Backdoor Attacks} introduce malicious triggers by modifying node attributes while keeping the graph structure unchanged. NFTA (Node Feature Target Attack) \citep{NFTA} injects feature triggers without requiring knowledge of GNN parameters, disrupting the feature space and confusing model predictions. It also introduces an adaptive strategy to balance feature smoothness. Xing et al. \citep{CGBA} selected trigger nodes with high similarity to neighbors, ensuring stealthiness without modifying labels or structure. However, both methods rely solely on feature manipulation, making them less effective when structural changes significantly impact message passing.

\textbf{Structure-based Backdoor Attacks} manipulate the graph topology by adding or removing edges to implant triggers. Zhang et al. \citep{zhang2021backdoor} introduced a subgraph-based trigger to mislead graph classification models while maintaining high attack success rates. Xi et al. \citep{GTA} extended this concept by generating adaptive subgraph triggers that dynamically tailor the attack for different inputs. Dai et al. \citep{UGBA} employed a bi-level optimization strategy to modify graph structures under an attack budget, maximizing stealthiness while ensuring effectiveness.

While homogeneous-centric methods struggle with heterogeneous structures, Chen et al. \cite{HBA} proposed HGBA, utilizing relation-based triggers via metapaths. However, HGBA relies on static trigger nodes and fixed connection patterns, lacking the adaptivity to diverse target features required for stealthy attacks.

\subsection{Attacks on Heterogeneous Graph Neural Networks}
Heterogeneous graphs have shown vulnerabilities under adversarial attacks, and several studies have explored this area. Zhang et al. \citep{RoHe} proposed RoHe, a robust HGNN framework that defends against adversarial attacks by pruning malicious neighbors using an attention purifier. Zhao et al. \citep{HGAttack} introduced HGAttack, the first grey-box evasion attack specifically targeting HGNNs, which leverages a semantic-aware mechanism and a novel surrogate model to generate perturbations. These works highlight the susceptibility of HGNNs to adversarial manipulations and the progress made in addressing these threats.

However, while adversarial attacks on heterogeneous graphs and backdoor attacks on homogeneous graphs have been explored, no prior work has investigated backdoor vulnerabilities in HGNNs. Our research addresses this gap by proposing a novel backdoor attack method specifically designed for heterogeneous graphs, leveraging their unique structural properties to embed triggers while maintaining high attack success rates and stealthiness.

\section{Preliminaries and problem formulation}
In this section, we introduce the preliminaries of backdoor attacks on heterogeneous graphs and define the problem. Table \ref{tab:notation} summarizes the notation used throughout this section for clarity.

\subsection{Preliminaries}
\textbf{Definition 3.1 (Heterogeneous graph)}
A heterogeneous graph is defined as $G=\{\mathcal{V}, \mathcal{E}, X\}$, where $\mathcal{V}=\left\{v_1, v_2, \ldots, v_n\right\}$ is the node set, and $X \in \mathbb{R}^{|\mathcal{V}| \times d}$ is a node feature matrix with $d$ being the dimension of each node feature.

The set $\mathcal{T}=\left\{t_1, t_2, \ldots, t_T\right\}$ represents $T$ different node types, where each node $v \in \mathcal{V}$ belongs to one specific type $t \in \mathcal{T}$. Nodes for each type $t$ is represented by the subset $\mathcal{V}_t$, and its size is denoted as $\left\|\mathcal{V}_t\right\|$. For each node type $t \in \mathcal{T}$, we denote $X_t$ as the feature set of all nodes of type $t$, and $x_t \in X_t$ represents the feature of an individual node of that type.
The set of edge types is denoted as
$
\mathcal{R}=\left\{r_{t_a, t_b} \mid t_a, t_b \in \mathcal{T}, t_a \neq t_b\right\}
$
where each edge type $r_{t_a, t_b}$ represents connections between nodes of type $t_a$ and nodes of type $t_b$. For each pair of node types $\left(t_a, t_b\right)$, we maintain an adjacency matrix $A_{t_a, t_b} \in\{0,1\}^{\left|\mathcal{V}_{t_a}\right| \times\left|\mathcal{V}_{t_b}\right|}$, where $A_{t_a, t_b}\left(v_i, v_j\right)=1$ indicates an edge between node $v_i \in \mathcal{V}_{t_a}$ and node $v_j \in \mathcal{V}_{t_b}$. We then define the edge set $\mathcal{E}$ as the union of all such edges, recorded as triples $\left(v_i, v_j, r_{t_a, t_b}\right)$ :
$
\mathcal{E}=\bigcup_{r_{t_a, t_b} \in \mathcal{R}}\left\{\left(v_i, v_j, r_{t_a, t_b}\right) \mid v_i \in \mathcal{V}_{t_a}, v_j \in \mathcal{V}_{t_b}, A_{t_a, t_b}\left(v_i, v_j\right)=1\right\}.
$
Hence, each adjacency matrix $A_{t_a, t_b}$ describes the connectivity between nodes of types $t_a$ and $t_b$, and each nonzero entry in $A_{t_a, t_b}$ corresponds to an edge in $\mathcal{E}$. A heterogeneous graph satisfies the condition $T+|\mathcal{R}|>2$.

\textbf{Definition 3.2 (Primary type, trigger type and auxiliary type)}
We define three key node-type categories in a heterogeneous graph.
The \emph{primary type} $t_p \in \mathcal{T}$ refers to the type of nodes on which the classification task is performed.
The \emph{trigger type} $t_{tr} \in \mathcal{T}$ denotes the type of nodes injected as backdoor triggers.
The \emph{auxiliary type} consists of all node types that are directly connected to
trigger-type nodes. Formally,
\[
\mathcal{T}_{\mathrm{aux}}
=
\left\{
t_a \in \mathcal{T}
\;\middle|\;
\exists\, 
v_{tr} \in \mathcal{V}_{t_{tr}},\,
v_a \in \mathcal{V}_{t_a} :
(v_{tr}, v_a, r_{t_{tr}, t_a}) \in \mathcal{E}
\right\}.
\]

\textbf{Definition 3.3 (Designated target class and non-target classes)}
Let $\mathcal{Y}$ denote the set of class labels in the classification task. The designated target class is defined as $y_t \in \mathcal{Y}$, representing the label to which the attacker aims to misclassify certain nodes. The non-target classes are given by $\mathcal{Y}_{\neg t}=\mathcal{Y} \backslash\left\{y_t\right\}$.
In a heterogeneous graph, classification is performed on nodes of the primary type $t_p$, whose node set is denoted as $\mathcal{V}_{t_p}$. Based on their ground-truth labels, we define:
$
\mathcal{V}_{y_t}=\left\{v \in \mathcal{V}_{t_p} \mid y_v=y_t\right\}, \quad \mathcal{V}_{\neg y_t}=\left\{v \in \mathcal{V}_{t_p} \mid y_v \neq y_t\right\}
$. By definition, $\mathcal{V}_{y_t} \cup \mathcal{V}_{\neg y_t}=\mathcal{V}_{t_p}$ and $\mathcal{V}_{y_t} \cap \mathcal{V}_{\neg y_t}=\emptyset$.

\begin{figure}[t]
    \centering
    \includegraphics[width=1.0\linewidth]{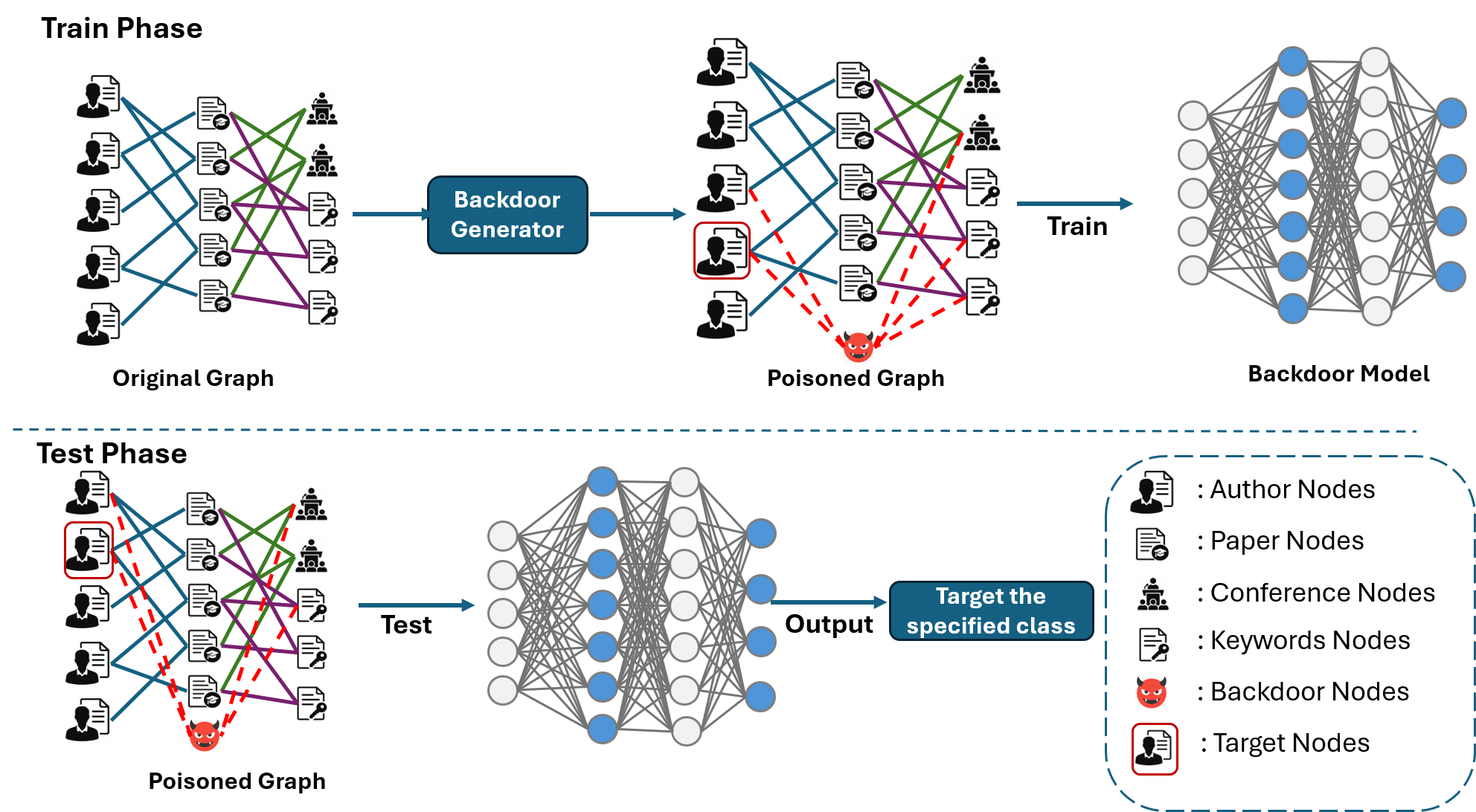}
    \caption{Overall Backdoor Attack Process on a Heterogeneous Graph.}
    \label{fig:wholeprocess}
\end{figure}

\subsection{Problem definition}
Given a heterogeneous graph $G=(\mathcal{V}, \mathcal{E}, X)$ and a node classification model $f_\theta: \mathcal{V}_{t_p} \rightarrow \mathcal{Y}$ trained on the primary node type $\mathcal{V}_{t_p} \subseteq \mathcal{V}$, a backdoor attack alters $G$ to construct a poisoned graph $\widetilde{G}=(\widetilde{\mathcal{V}}, \widetilde{\mathcal{E}}, \widetilde{X})$, ensuring that after training on $\widetilde{G}$, the model misclassifies specific target nodes while preserving overall classification accuracy. To construct $\widetilde{G}$, the attacker introduces a set of new trigger nodes $\mathcal{V}_{t_{\mathrm{tr}}}^{(\mathrm{new})}$ with feature matrix $X^{(\mathrm{new})}$ and new edges $\mathcal{E}^{(\mathrm{new})}$ connecting them to existing nodes. Specifically, $v^{(\mathrm{new})} \in \mathcal{V}_{t_{\mathrm{tr}}}^{(\mathrm{new})}$ denotes a newly injected trigger node, and $x^{(\mathrm{new})} \in \mathbb{R}^{d}$ denotes its feature vector; stacking the features of all new trigger nodes yields $X^{(\mathrm{new})} \in \mathbb{R}^{|\mathcal{V}_{t_{\mathrm{tr}}}^{(\mathrm{new})}| \times d}$. As a result, $\widetilde{\mathcal{V}}=\mathcal{V} \cup \mathcal{V}_{t_{\mathrm{tr}}}^{(\mathrm{new})}$, $\widetilde{\mathcal{E}}=\mathcal{E} \cup \mathcal{E}^{(\mathrm{new})}$, and $\widetilde{X}=\left[\begin{array}{c}X \\ X^{(\mathrm{new})}\end{array}\right]$. Specifically, the attacker selects a subset of primary-type nodes, denoted as $\mathcal{V}^{(p)} \subseteq \mathcal{V}_{t_p}$, as targeted victim nodes, and aims to enforce their wrongful classification into a designated target class $y_t$ during inference, i.e., $f_\theta(\widetilde{G}, v)=y_t, \forall v \in \mathcal{V}^{(p)}$. Meanwhile, for the remaining primary-type nodes $\mathcal{V}_{t_p} \backslash \mathcal{V}^{(p)}$, the model should retain its correct predictions, i.e., $f_\theta(\widetilde{G}, v)=y_v, \forall v \in \mathcal{V}_{t_p} \backslash \mathcal{V}^{(p)}$. Formally, the attack is formulated as an optimization problem:
\begin{equation}
\scriptsize 
\widetilde{G}^* =
\arg\max_{\widetilde{G} \,\in\, \mathcal{F}(G)}
\left[
    \sum_{v \in \mathcal{V}^{(p)}}
    \mathbf{1}\bigl(f_{\theta}(\widetilde{G}, v) = y_t\bigr)
    +
    \sum_{v \in \mathcal{V}_{t_p} \setminus \mathcal{V}^{(p)}}
    \mathbf{1}\bigl(f_{\theta}(\widetilde{G}, v) = y_v\bigr)
\right]
\end{equation}

Here, $\mathbf{1}(\cdot)$ is an indicator function that returns 1 if the condition is satisfied and 0 otherwise, and $\mathcal{F}(G)$ denotes the space of permissible modifications to $G$, which may include adding or modifying nodes, edges, or node features.

\begin{table}[t]
\small
\caption{Notation and Definitions}
\label{tab:notation}
\centering
\begin{tabular}{cl}
\toprule
Symbol & Meaning \\
\midrule
$G=(\mathcal{V},\mathcal{E},X)$ 
& Heterogeneous graph \\
$\mathcal{V}, \mathcal{E}, X$ 
& Nodes, edges, feature matrix \\
$X_t,\; x_t \in X_t$
& Type-$t$ features; single node feature \\
$\mathcal{T}, \mathcal{R}$ 
& Node/edge type sets \\
$t_p, t_{tr}$ 
& Primary/trigger node types \\
$\mathcal{T}_{\mathrm{aux}}$ 
& Auxiliary node types \\
$A_{t_a, t_b} \in\{0,1\}^{\left|\mathcal{V}_{t_a}\right| \times\left|\mathcal{V}_{t_b}\right|}$
& Adjacency matrix between types $t_a,t_b$ \\
$\mathcal{Y}$ 
& Class label set \\
$y_t$ 
& Target class \\
$\mathcal{V}_{t_p}, \mathcal{V}^{(p)}$
& Primary-type nodes, poisoned subset \\
$\mathcal{V}_{y_t}, \mathcal{V}_{\neg y_t}$ 
& Primary-type nodes w/ or w/o label $y_t$ \\
$\mathcal{V}_{t_{\mathrm{tr}}}^{(\mathrm{new})}$ 
& Newly added trigger nodes \\
$v^{(\mathrm{new})}$
& A newly added node (typically a trigger node) \\
$X^{(\mathrm{new})}, \mathcal{E}^{(\mathrm{new})}$
& New trigger-node features, edges \\
$\widetilde{G}=(\widetilde{\mathcal{V}}, \widetilde{\mathcal{E}}, \widetilde{X})$ 
& Poisoned graph \\
$f_{\theta}$ 
& Classification model \\
$\mathcal{F}(G)$ 
& Allowed modifications \\
$\mathbf{1}(\cdot)$ 
& Indicator function \\
\bottomrule
\end{tabular}
\end{table}

\section{Methodology}
\begin{figure*}
    \centering
    \includegraphics[width=1.0\linewidth]{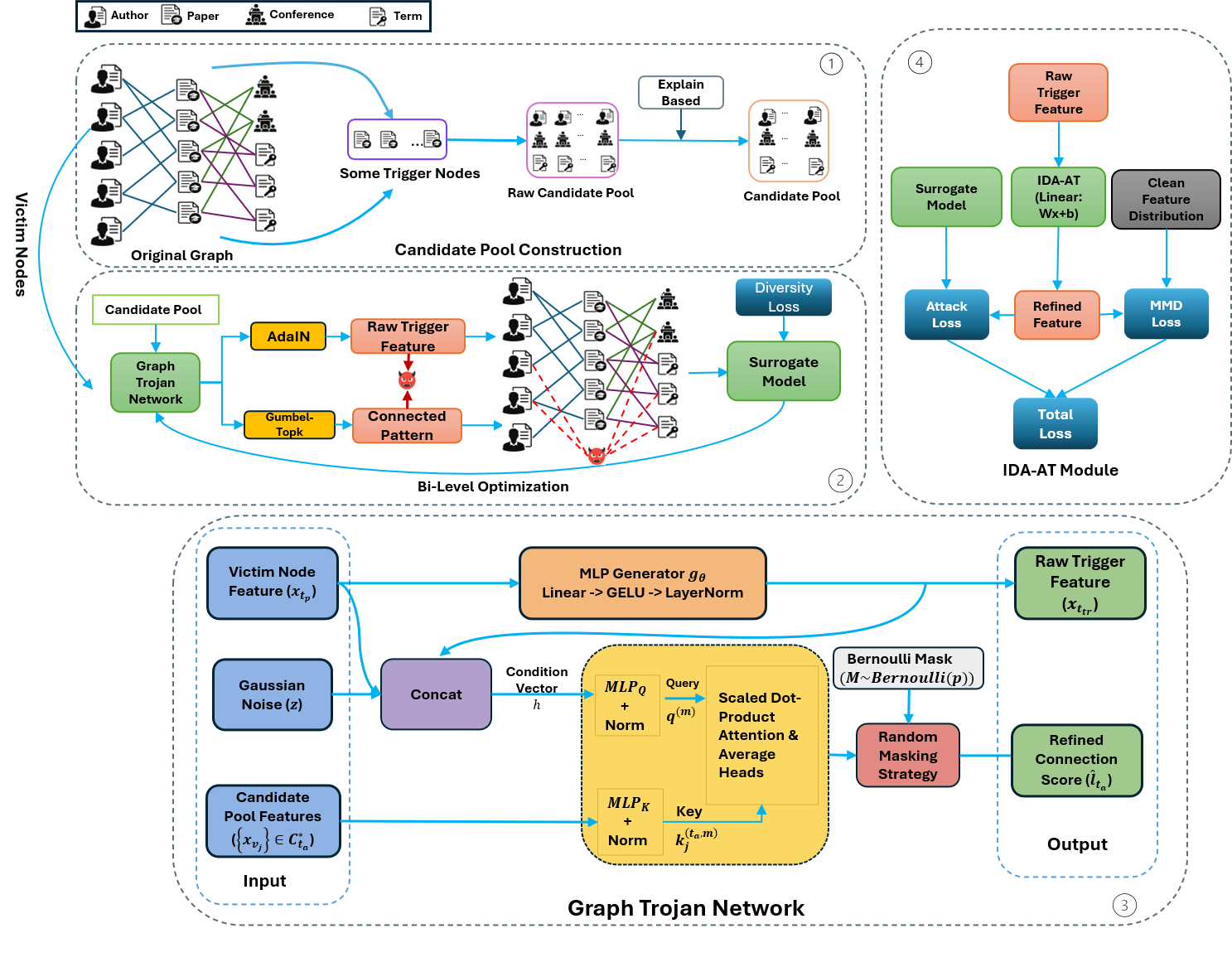}
    \caption{HeteroHBA Algorithm Description}
    \label{fig:HeteroHBA Algorithm}
\end{figure*}

In this section, we present the details of HeteroHBA, a framework designed to generate stealthy and effective backdoor attacks on heterogeneous graphs. Unlike traditional methods that optimize static triggers, HeteroHBA learns a generative mechanism to dynamically generate trigger nodes that are both structurally diverse and feature-consistent with the target domain.The framework is decomposed into three integrated stages. First, the Candidate Pool Construction module filters a refined set of potential auxiliary nodes using an explanation-based screening method. Following this, we employ GraphTrojanNet as the core generative module to produce trigger nodes with stealthy features via Adaptive Instance Normalization (AdaIN) and diverse structural patterns. Finally, we formulate the learning process as a bi-level optimization problem to maximize attack effectiveness. The overall workflow of HeteroHBA is conceptually illustrated in Fig.~\ref{fig:HeteroHBA Algorithm}. By integrating these components, we generate a poisoned graph $\widetilde{G} = (\widetilde{\mathcal{V}}, \widetilde{\mathcal{E}}, \widetilde{X})$ that achieves high attack success rates while preserving high stealthiness.

The pseudocode for the training procedure is provided in Appendix~\ref{appendix_sec: pseudocode}, with a complexity analysis in Appendix~\ref{appendix_sec:time complexity analysis}.

\subsection{Candidate Pool Construction}
As highlighted in Fig.~\ref{fig:HeteroHBA Algorithm} (\textcircled{1}), this stage constructs a refined auxiliary-node candidate pool that the generator will later connect triggers to.
We begin by identifying the primary-type nodes that belong to the target class:
\begin{equation}
\mathcal{V}_{y_t}
=\left\{ v \in \mathcal{V}_{t_p} ~\middle|~ y_v = y_t \right\}.
\end{equation}

For each target-class node $v_{t_p} \in \mathcal{V}_{y_t}$, we directly collect its 
trigger-mediated 2-hop auxiliary neighbors of types 
$t_a \in \mathcal{T}_{\mathrm{aux}}$. These neighbors are defined as:
\begin{equation}
\label{equ:auxiliary_node_calculation}
\begin{aligned}
\mathcal{V}_{t_a}(v_{t_p}) = \Big\{ & v_{t_a} \in \mathcal{V}_{t_a} ~\Big|~ \exists\, v_{t_{tr}} \in \mathcal{V}_{t_{tr}}, \\
& A_{t_p,t_{tr}}(v_{t_p}, v_{t_{tr}})=1,
~ A_{t_{tr},t_a}(v_{t_{tr}}, v_{t_a})=1 \Big\}.
\end{aligned}
\end{equation}

Aggregating across all target-class primary nodes yields the 
raw candidate pool for each auxiliary type:
\begin{equation}
\label{equ:raw_candidate_pool_calculation}
\mathcal{C}_{t_a}
=
\bigcup_{v_{t_p} \in \mathcal{V}_{y_t}}
\mathcal{V}_{t_a}(v_{t_p}),
\qquad
\forall~ t_a \in \mathcal{T}_{\mathrm{aux}}.
\end{equation}

The set $\mathcal{C}_{t_a}$ contains all auxiliary-type nodes reachable from 
target-class primary nodes via exactly two hops through a trigger-type node, 
forming the first stage of our candidate pool construction.
\begin{figure}[t]
    \centering
    
    \begin{subfigure}[b]{0.48\linewidth}
        \centering
        \includegraphics[width=\linewidth]{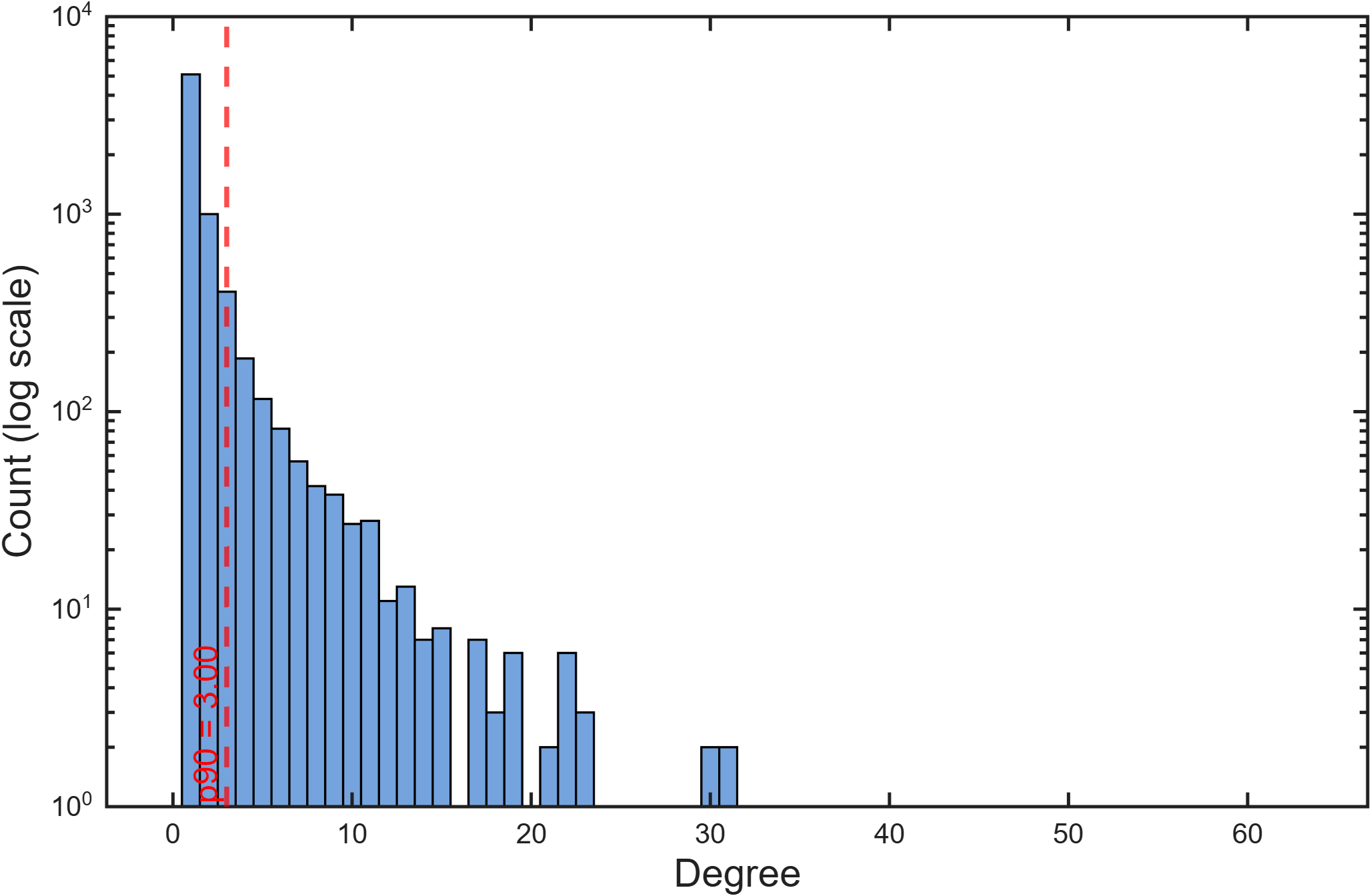}
        \caption{ACM Author Degree Distribution}
        \label{fig:acm_author}
    \end{subfigure}
    \hfill
    \begin{subfigure}[b]{0.48\linewidth}
        \centering
        \includegraphics[width=\linewidth]{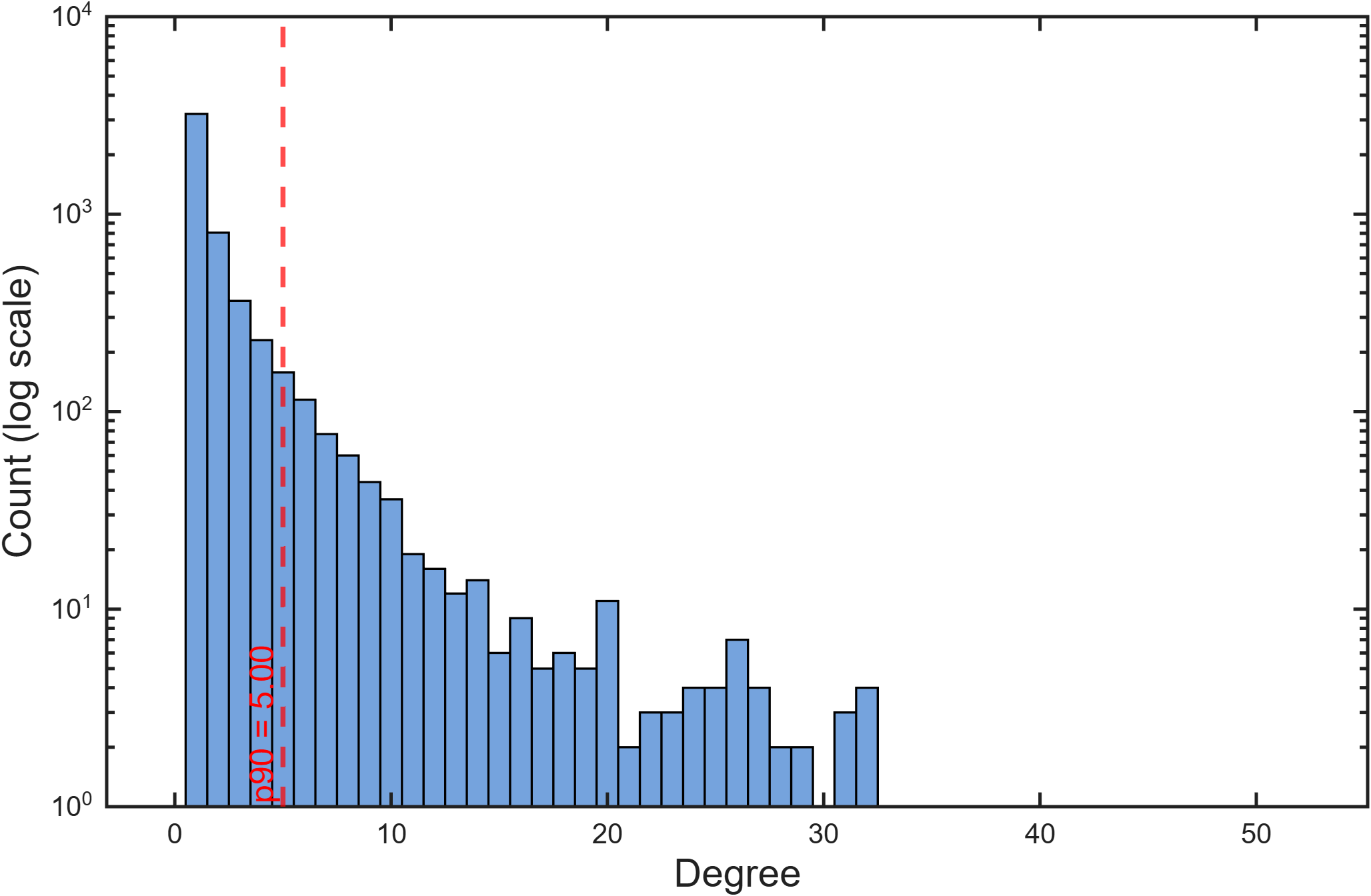}
        \caption{IMDB Actor Degree Distribution}
        \label{fig:acm_paper}
    \end{subfigure}

    \caption{Degree distributions of different node types.}
    \label{fig:degree_distribution}
\end{figure}
Having established the raw candidate pool, we next identify an appropriate pool size to balance attack effectiveness with computational efficiency. To determine the size of the candidate pool for each auxiliary type, we analyze the degree distribution of trigger-type nodes toward auxiliary nodes. As illustrated in Fig.~\ref{fig:degree_distribution}, real heterogeneous graphs exhibit clear long-tailed degree patterns: most nodes have very small degrees, while a small fraction exhibit much larger degrees.
In such distributions, the mean is sensitive to extreme values. 
Moreover, selecting excessively high degrees would make the generated trigger nodes stand out as high-degree hubs, increasing the risk of detection.
Therefore, to achieve strong attack effectiveness while maintaining stealthiness and following standard practices in robust statistics~\cite{huber2011robust}, we adopt the 90th percentile (P90) of the degree distribution as a robust estimate of the typical upper bound on the degree of generated trigger nodes between trigger and auxiliary types. 
For each auxiliary type $t_a$, we compute:
\begin{equation}
\label{equ:candidate_pool_size1}
K_{t_a}
=
\left\lceil 
\mathrm{Quantile}_{0.9}
\left(
\left\{
d_{t_{tr}\rightarrow t_a}(v) \mid v\in\mathcal{V}_{t_{tr}}
\right\}
\right)
\right\rceil .
\end{equation}
where $d_{t_{tr} \to t_a}(v)$ represents the degree (number of connections) of the trigger-type node $v$ to the auxiliary-type node $t_a$. 

Then to provide sufficient search space for downstream optimization, we further enlarge this base size by a hyperparameter $n$, resulting in the final candidate pool size:
\begin{equation}
\label{equ:candidate_pool_size2}
K^{(\mathrm{pool})}_{t_a} = n \cdot K_{t_a}.
\end{equation}

Having defined the target capacity for the final candidate pool, we must now select the $K^{(\mathrm{pool})}_{t_a}$ most influential nodes from the raw candidate pool based on their actual impact. We adopt a saliency-based selection strategy \cite{saliencymap}, to determine which nodes from the raw candidate pool should be retained in the final candidate set. 
For each auxiliary type $t_a \in \mathcal{T}_{\mathrm{aux}}$, we refine the raw candidate pool $\mathcal{C}_{t_a}$ using a saliency-based importance scoring mechanism \cite{saliencyscore}. 
We first train a heterogeneous graph classifier as a surrogate model on the clean graph and identify all correctly predicted target-class primary nodes 
$v_{t_p} \in \mathcal{V}_{y_t}$.
For each such node, we compute the saliency score $\mathrm{Sal}(v_{t_a}, v_{t_p})$ for every candidate node $v_{t_a} \in \mathcal{C}_{t_a}$, which measures how influential $v_{t_a}$ is to the prediction results by quantifying the sensitivity of the model output to this node.

We then aggregate the saliency scores across all target-class primary nodes to obtain a global importance value:

\begin{equation}
\label{equ:saliency_calculation}
S(v_{t_a}) = \sum_{v_{t_p} \in \mathcal{V}_{y_t}} \mathrm{Sal}(v_{t_a}, v_{t_p}), \qquad v_{t_a} \in \mathcal{C}_{t_a}.
\end{equation}

Finally, we rank all nodes in $\mathcal{C}_{t_a}$ according to $S(v_{t_a})$
in descending order, and select the top 
$K^{(\mathrm{pool})}_{t_a}$ nodes as the final filtered candidate set $C^*_{t_a}$ for type $t_a$.

\subsection{GraphTrojanNet: Learning Trigger Features and Connection Patterns}
GraphTrojanNet serves as the core generative module of our framework, aiming to dynamically synthesize feature representations for new nodes as well as their connection patterns with auxiliary candidate nodes, as shown in Fig.~\ref{fig:HeteroHBA Algorithm} (\textcircled{3}). Specifically, we adopt a Multi-Layer Perceptron (MLP)--based generator $g_{\theta}$, which consists of stacked linear transformations, GELU activation functions~\cite{gelu}, and layer normalization layers. Given a victim node with original feature vector $x_{t_p}$, the generator maps it into the trigger feature space as:
\begin{equation}
x^{(\mathrm{new})} = g_{\theta}(x_{t_p}).
\end{equation}
This non-linear transformation produces victim-aware trigger features, making the generated triggers more effective for influencing the target node.

To model the connection topology of the trigger node, we employ a multi-head attention mechanism conditioned on a composite condition vector $h$. The condition vector is constructed by concatenating the victim node features $x_{t_p}$, the generated trigger f-deterministic connecteatures $x^{(\mathrm{new})}$, and a Gaussian noise vector $z$, which introduces stochasticity to promote diverse and nonion patterns~\cite{cGAN}:
\begin{equation}
h = [x_{t_p} \parallel x^{\mathrm(\mathrm{new})} \parallel z].
\end{equation}

Based on the condition vector $h$, queries are generated and matched against keys projected from the auxiliary candidate node pool $C^*_{t_a}$ to estimate the likelihood of forming connections. The connection score (logit) between the trigger node and a candidate node $v_j \in C^*_{t_a}$ is computed as:
\begin{equation}
l_{t_a}(j) = \frac{1}{H} \sum_{m=1}^{H} \frac{\langle q^{(m)}, k_j^{(t_a, m)} \rangle}{\sqrt{d_m}},
\end{equation}
where
\begin{equation}
q^{(m)} = \operatorname{Norm}\big([\operatorname{MLP}_Q(h)]_m\big)
\end{equation}
denotes the query vector derived from the condition vector for the $m$-th attention head, and
\begin{equation}
k_j^{(t_a, m)} = \operatorname{Norm}\big([\operatorname{MLP}_K(x_{v_j})]_m\big)
\end{equation}
represents the corresponding key vector generated from the feature representation $x_{v_j}$ of candidate node $v_j$.

Crucially, to prevent mode collapse, where the generator converges to a single and static connection pattern regardless of the input, we implement a random masking strategy \cite{GAN}. By stochastically suppressing a subset of connection scores with a Bernoulli mask $M$, we force the network to explore diverse structural configurations. This ensures that the generated backdoors are distinct and tailored to the specific characteristics of each victim node, rather than collapsing into a same pattern.
\begin{equation}
\hat{l}_{t_a} = \text{Mask}(l_{t_a}, M), \quad \text{where } M \sim \text{Bernoulli}(p_{mask})
\end{equation}

Finally, GraphTrojanNet outputs these refined connection scores $\hat{l}_{t_a}$ alongside the raw generated trigger features $x_{t_{tr}}$ for subsequent structure sampling and model optimization.

\subsection{Bi-Level Optimization Framework}
In this subsection, we present the proposed Bi-Level Optimization Framework (see Fig.~\ref{fig:HeteroHBA Algorithm}, \textcircled{2}) designed to generate stealthy and effective backdoor triggers. Specifically, the trigger generation process is decomposed into two coupled components: generating node features that preserve distributional consistency with the target domain, and identifying optimal connection patterns to auxiliary nodes. To jointly optimize these components, we formulate the problem as a bi-level optimization task. The upper level optimizes the generator to maximize attack success rate, while the lower level updates a surrogate model to simulate the victim's response to the poisoned data. The detailed methodologies for feature generation, connection learning, and the overall optimization strategy are described below.

\textbf{Trigger Feature Generation} To ensure the synthesized features blend seamlessly with the target environment, we employ the Adaptive Instance Normalization (AdaIN) mechanism \cite{adaIN}. Specifically, for each injected trigger node, its raw generated feature $x_{t_{tr}}$ is transformed to align its statistics with the clean data distribution. This is achieved by first standardizing the raw feature using batch statistics and then scaling and shifting it using the global mean ($\mu_{\text{clean}}$) and standard deviation ($\sigma_{\text{clean}}$) computed from all clean trigger nodes $\mathcal{V}_{t_{tr}}$. The adapted trigger feature $\hat{x}^{(\mathrm{new})}$ is formally computed as:
\begin{equation}
  \small
  \label{equ:feature_generation}
\hat{x}^{(\mathrm{new})} = \text{AdaIN}(x^{(\mathrm{new})}, \mathcal{V}_{t_{tr}}) = \sigma_{\text{clean}} \left( \frac{x^{(\mathrm{new})} - \mu(x^{\mathrm{new}})}{\sigma(x^{\mathrm{new}})} \right) + \mu_{\text{clean}}
\end{equation}
where $\mu(x^{(\mathrm{new})})$ and $\sigma(x^{(\mathrm{new})})$ are the mean and standard deviation of the generated features within the current batch (corresponding to a subset of $\mathcal{V}^{(\text{new})}$), and $\mu_{\text{clean}}$ and $\sigma_{\text{clean}}$ are the pre-computed global mean and standard deviation. This adaptation forces the synthesized features to instantly conform to the target feature space, thereby significantly enhancing stealthiness and preventing detectable feature anomalies.

\textbf{Connection pattern generation} Given the connection logits, we construct $\mathcal{E}^{(\text{new})}$ by selecting, for each trigger node and each auxiliary type $t_a$, the top-$K_{t_a}$ candidates from $C_{t_a}^*$.
Since the discrete Top-$k$ operator is non-differentiable, we employ a Gumbel Top-$k$ relaxation with the Straight-Through Estimator~\cite{ahle2022differentiable,straightthrough} to enable end-to-end training while keeping the forward pass discrete.
Formally,
\begin{equation}
\label{equ:new_edges_def}
\mathcal{E}^{(\text{new})}\!\left(v^{(\mathrm{new})}\right)
=
\bigcup_{t_a \in \mathcal{T}_{\mathrm{aux}}}
\left\{
\left(v^{(\mathrm{new})},\, u,\, r_{t_{tr}, t_a}\right)
\;\middle|\;
u \in \operatorname{Top}\text{-}k\bigl(\hat{l}_{t_a},\, K_{t_a}\bigr)
\right\}.
\end{equation}
where $\hat{l}_{t_a}^{(i)}$ denotes the connection score vector over $C_{t_a}^*$ after masking/Gumbel relaxation. The detailed derivation is provided in Appendix~\ref{appendix:topk_proof}.

To further prevent the generator from collapsing into identical edge patterns across different triggers, while ensuring that the diversity loss does not interfere with training at an early stage, we introduce a simple diversity loss inspired by the hinge loss objective~\cite{hingemargin}. Let $p_i \in \{0,1\}^C$ denote the binary Top-$k$ selection vector for the $i$-th trigger, and let $\tilde{p}_i = p_i / \|p_i\|_2$ be its normalized form. The loss is defined as:
\begin{equation}
  \label{equ:diversity_loss}
\mathcal{L}_{\text{div}}
=
\frac{1}{B(B-1)}
\sum_{\substack{i\neq j}}
\max\left(0,\, \tilde{p}_i^\top \tilde{p}_j - \tau\right),
\end{equation}
where $B$ is the batch size and $\tau$ is a similarity margin. This loss penalizes overly similar connection patterns, thereby encouraging diverse and less detectable backdoor structures.


\textbf{Bi-level optimization}
To jointly optimize the surrogate model and the trigger generator, we adopt a bi-level optimization framework. 
Given a fixed generator parameter $\theta_g$, the generated trigger features and selected auxiliary-node connections are injected into the clean graph $G$ to construct the poisoned training graph $\tilde G(\theta_g)$. 
The surrogate model $f_s(\cdot;\theta_s)$ is then trained on $\tilde G(\theta_g)$ by minimizing the following lower-level objective:
\begin{equation}
\begin{aligned}
\min_{\theta_s}\; \mathcal{L}_s(\theta_s,\theta_g)
= &\;
\sum_{v_i \in V_{t_p} \setminus V^{(p)}}
\ell\!\left(f_s(G, v_i;\theta_s), y_i\right) \\
& +
\sum_{v_i \in V^{(p)}}
\ell\!\left(f_s(\tilde G(\theta_g), v_i;\theta_s), y_t\right),
\end{aligned}
\label{eq:inner}
\end{equation}
where the first term preserves the surrogate model's normal predictive behavior, and the second term optimizes the backdoor objective on poisoned nodes.  

In the upper level, we fix the surrogate model parameters $\theta_s$ trained in the lower-level optimization and optimize the trigger generator such that attaching the generated triggers induces the surrogate model to predict the target class $y_t$.
We attach triggers to a broader node set and obtain the outer-level poisoned graph $\hat G(\theta_g)$. 
The attack objective is defined as
\begin{equation}
  \label{equ:classification_loss}
\mathcal{L}_g(\theta_s,\theta_g)
=
\sum_{v_i \in V_{t_p}}
\ell\!\left(f_s(\hat G(\theta_g), v_i;\theta_s),\, y_t\right).
\end{equation}

We further incorporate the diversity loss $\mathcal{L}_{\mathrm{div}}$, resulting in the following bi-level optimization problem:
\begin{equation}
\begin{aligned}
\min_{\theta_g} \quad & \mathcal{L}_g(\theta_s^*(\theta_g), \theta_g) + \lambda_{\text{div}} \mathcal{L}_{\text{div}}(\theta_g) \\
\text{s.t.} \quad & \theta_s^*(\theta_g) = \arg\min_{\theta_s} \mathcal{L}_s(\theta_s, \theta_g).
\end{aligned}
\label{eq:bilevel_final}
\end{equation}

\subsection{Optimization Algorithm}
We use an alternating optimization strategy to solve the bi-level
objective in Eq.~(\ref{eq:bilevel_final}). In each outer iteration, we
first update the surrogate model and then update the trigger
generator.

\textbf{Updating the Lower-Level Surrogate Model.}
Computing $\theta_s^\ast(\theta_g)$ exactly is costly. To make the
procedure practical, we approximate it by running $N$ gradient steps
on $\mathcal{L}_s(\theta_s,\theta_g)$ while keeping $\theta_g$ fixed:
\begin{equation}
  \label{equ:low_level_optimization}
\theta_s^{t+1}
=
\theta_s^{t}
-
\alpha_s \nabla_{\theta_s}
\mathcal{L}_s(\theta_s^{t},\theta_g),
\end{equation}
where $\theta_s^{t}$ is the parameter value at iteration $t$, and
$\alpha_s$ is the learning rate for the surrogate model.

\textbf{Updating the Upper-Level Trigger Generator.}
Once the $N$ updates of the surrogate model are finished, the obtained
$\theta_s^{T}$ is used in the upper-level objective. To avoid the high
cost of full bi-level differentiation, we adopt a first-order update
for $\theta_g$:
\begin{equation}
  \label{equ:upper_level_optimization}
\theta_g^{k+1}
=
\theta_g^{k}
-
\alpha_g
\nabla_{\theta_g}
\Big(
\mathcal{L}_g(\bar{\theta}_s, \theta_g^{k})
+
\lambda_{\mathrm{div}} L_{\mathrm{div}}(\theta_g^{k})
\Big),
\end{equation}
where $\bar{\theta}_s$ indicates that gradients are not propagated
through the surrogate model, and $\alpha_g$ is the learning rate for
the trigger generator.

\textbf{Overall Procedure.}
By repeatedly applying Eqs.~(\ref{equ:low_level_optimization}) and
(\ref{equ:upper_level_optimization}), the surrogate model and the trigger generator
are updated in turn. This alternating process allows the trigger
generator to gradually learn trigger patterns that achieve the desired
target outputs while satisfying the regularization terms. After the
training finishes, we use the final generator to produce triggers and
construct the poisoned graph. We then perform a lightweight
post-generative refinement step to further improve the statistical
stealthiness of the generated features.

\subsection{Post-Generative Refinement via IDA-AT}
\label{sec:post_refinement}
Although AdaIN aligns the first and second-order statistics during trigger generation (Eq.~\ref{equ:feature_generation}), it does not explicitly constrain higher-order distribution information, which may leave detectable statistical anomalies. To address this limitation after the bi-level optimization converges, we introduce an Invertible Distribution-Aligned Affine Transformation (IDA-AT) module as a post-processing unit for the generator (parameterized by $\theta_g$) to refine the feature distribution, as shown in Fig.~\ref{fig:HeteroHBA Algorithm}~(\textcircled{4}).

This module is implemented as a linear layer without activation functions:
\begin{equation}
x^{(\mathrm{new})}_{aff} = W \hat{x}^{(\mathrm{new})} + b.
\end{equation}
The removal of non-linearities ensures the invertibility of the transformation, theoretically enabling the victim model $f_\theta$ to capture the inverse mapping and recover the adversarial semantics learned through bi-level optimization.

To enforce rigorous statistical stealthiness, we minimize the Maximum Mean Discrepancy (MMD) loss \cite{mmdloss}:
\begin{equation}
  \label{eq:ida_at_mmd_loss}
\begin{aligned}
  \mathcal{L}_{MMD} = & \frac{1}{n^2} \sum_{u \in X_{aff}} \sum_{v \in X_{aff}} k(u, v) + \frac{1}{m^2} \sum_{u \in X^{(\mathrm{new})}} \sum_{v \in X^{(\mathrm{new})}} k(u, v) \\
  & - \frac{2}{nm} \sum_{u \in X_{aff}} \sum_{v \in X^{(\mathrm{new})}} k(u, v)
\end{aligned}
\end{equation}

In this formulation, $k(\cdot, \cdot)$ represents a kernel function (e.g., Gaussian RBF) that implicitly maps the features into a high-dimensional Reproducing Kernel Hilbert Space (RKHS), allowing the model to minimize the divergence between poisoned and clean distributions across all statistical moments.

Concurrently, we utilize the surrogate model $f_s$ obtained from the bi-level optimization to compute the attack alignment loss, defined as:
\begin{equation}
\label{eq:ida_at_atttack_loss}
\mathcal{L}_{atk-aff} = \sum_{v_i \in \mathcal{V}^{(p)}} \ell(f_s(\tilde{G}(x_{aff}), v_i; \theta_s), y_t),
\end{equation}
which ensures the affine transformation preserves the trigger's potency in misleading the target nodes $\mathcal{V}^{(p)}$.

Ultimately, the total loss is formulated as:
\begin{equation}
\label{eq:IDA-AT loss}
\mathcal{L} = \mathcal{L}_{MMD} + \mathcal{L}_{atk-aff},
\end{equation}
to jointly control the training of the post-processing unit. This dual-objective optimization empowers HeteroHBA to bypass distribution-sensitive defenses while maintaining high attack success rates. This post-processing stage is performed after the bi-level optimization converges, ensuring that the refined triggers achieve rigorous statistical stealthiness while preserving the adversarial semantics learned through bi-level optimization.

\section{Experiments}
In this section, we evaluate our proposed method on multiple benchmark datasets to investigate the following research questions:
\begin{enumerate}[label=\textbf{RQ\arabic*:}]
    \item \textbf{Attack Effectiveness:} How effective is the proposed HeteroHBA compared to state-of-the-art baseline attacks on various datasets?
    \item \textbf{Defense Resistance:} Is the proposed attack still effective against potential defense mechanisms?
    \item \textbf{Hyperparameter Sensitivity:} How do different hyperparameters (e.g., random mask rate, ) affect the attack performance?
    \item \textbf{Ablation Study:} What is the contribution of each component (e.g., influential node selection, adaIN mechanism) to the overall success of the attack?
\end{enumerate}

\subsection{Experimental Settings}
\subsubsection{Datasets}
We evaluate our method on three real-world heterogeneous datasets: DBLP, ACM, and IMDB\citep{MAGNN}. DBLP consists of four entity types (authors, papers, terms, conferences), with authors being categorized into three research areas (database, data mining, artificial intelligence). ACM includes papers from KDD, SIGMOD, SIGCOMM, MobiCOMM, and VLDB, being categorized into three fields (database, wireless communication, data mining). IMDB contains movies, keywords, actors, and directors, with movies being classified into action, comedy, and drama. The statistics of these datasets are shown in Table~\ref{tab:dataset statistics}.

\begin{table}[ht]
\centering
\small
\caption{Dataset Statistics}
\label{tab:dataset statistics}
\resizebox{\linewidth}{!}{
\begin{tabular}{lccccc}
\toprule
\textbf{Dataset} & \textbf{\#Node Types} & \textbf{\#Edge Types} & \textbf{\#Nodes} & \textbf{\#Edges} & \textbf{Primary Type} \\
\midrule
ACM   & 3 & 4 & 11252  & 34864  & paper  \\
IMDB  & 3 & 4 & 11616  & 34212  & movie \\
DBLP  & 4 & 6 & 26198  & 242142 & author  \\
\bottomrule
\end{tabular}%
}
\end{table}

\begin{table*}[h]
\centering
\caption{Comparison of Attack Performance (CAD and ASR). For both metrics, the best result is \textbf{bold} and the second best is \textit{italic}. Note that for CAD, lower is better, while for ASR, higher is better. Standard deviations $> 0.1$ are marked with $^{\dagger}$.}
\label{tab:attack_performance}
\resizebox{\textwidth}{!}{%
\begin{tabular}{ccccccccccc}
\toprule
\multirow{2}{*}{Dataset} & \multirow{2}{*}{Model} & \multirow{2}{*}{Class} & \multicolumn{2}{c}{CGBA} & \multicolumn{2}{c}{HGBA} & \multicolumn{2}{c}{HeteroHBA-Variant $\mathrm{I}$} & \multicolumn{2}{c}{HeteroHBA (Ours)} \\
\cmidrule(lr){4-5} \cmidrule(lr){6-7} \cmidrule(lr){8-9} \cmidrule(lr){10-11}
 &  &  & CAD & ASR & CAD & ASR & CAD & ASR & CAD & ASR \\ 
\midrule

\multirow{9}{*}{ACM} & \multirow{3}{*}{HAN} & 0 & $0.0036_{\pm 0.1340^{\dagger}}$ & $0.1721_{\pm 0.1623^{\dagger}}$ & $\mathbf{-0.0692}_{\pm 0.1093^{\dagger}}$ & $0.0348_{\pm 0.0336}$ & $0.1203_{\pm 0.0806}$ & $0.0962_{\pm 0.0029}$ & $\mathit{-0.0568}_{\pm 0.0532}$ & $\mathbf{0.9867}_{\pm 0.0188}$ \\
 &  & 1 & $0.0017_{\pm 0.1229^{\dagger}}$ & $0.3672_{\pm 0.0792}$ & $\mathbf{-0.0242}_{\pm 0.0565}$ & $0.4100_{\pm 0.3504^{\dagger}}$ & $0.0298_{\pm 0.1002^{\dagger}}$ & $\mathit{0.7944}_{\pm 0.3562^{\dagger}}$ & $0.0127_{\pm 0.0063}$ & $\mathbf{0.9983}_{\pm 0.0029}$ \\
 &  & 2 & $\mathbf{-0.0232}_{\pm 0.1273^{\dagger}}$ & $0.1791_{\pm 0.1098^{\dagger}}$ & $0.0212_{\pm 0.1187^{\dagger}}$ & $0.0149_{\pm 0.0061}$ & $0.0292_{\pm 0.1168^{\dagger}}$ & $0.0083_{\pm 0.0144}$ & $\mathit{-0.0226}_{\pm 0.0163}$ & $\mathbf{0.9917}_{\pm 0.0057}$ \\
 \cmidrule(lr){2-11}
 & \multirow{3}{*}{HGT} & 0 & $0.0152_{\pm 0.0151}$ & $0.3721_{\pm 0.1695^{\dagger}}$ & $\mathit{-0.0026}_{\pm 0.0134}$ & $\mathit{0.9960}_{\pm 0.0089}$ & $\mathbf{-0.0044}_{\pm 0.0183}$ & $\mathbf{0.9983}_{\pm 0.0029}$ & $0.0155_{\pm 0.0069}$ & $\mathbf{0.9701_{\pm 0.0348}}$ \\
 &  & 1 & $0.0132_{\pm 0.0129}$ & $0.4955_{\pm 0.2912^{\dagger}}$ & $0.0079_{\pm 0.0159}$ & $0.9930_{\pm 0.0156}$ & $\mathit{0.0022}_{\pm 0.0051}$ & $\mathbf{1.0000}_{\pm 0.0000}$ & $\mathbf{0.0017}_{\pm 0.0088}$ & $\mathit{0.9950}_{\pm 0.0050}$ \\
 &  & 2 & $0.0093_{\pm 0.0054}$ & $0.4388_{\pm 0.2079^{\dagger}}$ & $\mathit{0.0056}_{\pm 0.0232}$ & $0.8149_{\pm 0.4083^{\dagger}}$ & $\mathbf{-0.0083}_{\pm 0.0184}$ & $\mathbf{1.0000}_{\pm 0.0000}$ & $0.0077_{\pm 0.0110}$ & $\mathit{0.9900}_{\pm 0.0100}$ \\
 \cmidrule(lr){2-11}
 & \multirow{3}{*}{SimpleHGN} & 0 & $0.0093_{\pm 0.0117}$ & $0.6408_{\pm 0.1162^{\dagger}}$ & $\mathit{-0.0007}_{\pm 0.0137}$ & $0.3502_{\pm 0.4258^{\dagger}}$ & $\mathbf{-0.0088}_{\pm 0.0096}$ & $\mathbf{1.0000}_{\pm 0.0000}$ & $-0.0099_{\pm 0.0060}$ & $\mathit{0.9751}_{\pm 0.0199}$ \\
 &  & 1 & $0.0056_{\pm 0.0087}$ & $0.4299_{\pm 0.2485^{\dagger}}$ & $\mathbf{-0.0050}_{\pm 0.0037}$ & $0.8348_{\pm 0.2555^{\dagger}}$ & $0.0077_{\pm 0.0019}$ & $\mathbf{1.0000}_{\pm 0.0000}$ & $\mathit{-0.0044}_{\pm 0.0124}$ & $\mathit{0.9685}_{\pm 0.0546}$ \\
 &  & 2 & $0.0066_{\pm 0.0176}$ & $0.5075_{\pm 0.1282^{\dagger}}$ & $0.0070_{\pm 0.0134}$ & $0.4070_{\pm 0.5391^{\dagger}}$ & $\mathit{0.0011}_{\pm 0.0120}$ & $\mathbf{1.0000}_{\pm 0.0000}$ & $\mathbf{-0.0094}_{\pm 0.0034}$ & $\mathit{0.9801}_{\pm 0.0263}$ \\
\midrule

\multirow{9}{*}{DBLP} & \multirow{3}{*}{HAN} & 0 & $\mathit{0.0125}_{\pm 0.0100}$ & $0.3291_{\pm 0.2261^{\dagger}}$ & $0.0141_{\pm 0.0508}$ & $0.0837_{\pm 0.0184}$ & $\mathbf{-0.0181}_{\pm 0.0328}$ & $\mathit{0.9984}_{\pm 0.0028}$ & $0.0384_{\pm 0.0468}$ & $\mathbf{1.0000}_{\pm 0.0000}$ \\
 &  & 1 & $0.0122_{\pm 0.0043}$ & $0.1133_{\pm 0.0945}$ & $0.0125_{\pm 0.0211}$ & $\mathit{0.7517}_{\pm 0.1556^{\dagger}}$ & $\mathbf{0.0115}_{\pm 0.0300}$ & $0.7126_{\pm 0.4977^{\dagger}}$ & $0.0186_{\pm 0.0038}$ & $\mathbf{1.0000}_{\pm 0.0000}$ \\
 &  & 2 & $0.0010_{\pm 0.0225}$ & $0.1744_{\pm 0.1501^{\dagger}}$ & $\mathit{-0.0046}_{\pm 0.0138}$ & $0.0443_{\pm 0.0338}$ & $\mathbf{-0.1124}_{\pm 0.1706^{\dagger}}$ & $\mathit{0.4844}_{\pm 0.4719^{\dagger}}$ & $0.0049_{\pm 0.0059}$ & $\mathbf{1.0000}_{\pm 0.0000}$ \\
 \cmidrule(lr){2-11}
 & \multirow{3}{*}{HGT} & 0 & $0.0135_{\pm 0.0105}$ & $0.5310_{\pm 0.0368}$ & $0.0053_{\pm 0.0102}$ & $0.0404_{\pm 0.0530}$ & $\mathbf{0.0016}_{\pm 0.0033}$ & $\mathbf{1.0000}_{\pm 0.0000}$ & $0.0077_{\pm 0.0019}$ & $\mathit{0.9885}_{\pm 0.0124}$ \\
 &  & 1 & $0.0125_{\pm 0.0054}$ & $0.3783_{\pm 0.1832^{\dagger}}$ & $0.0141_{\pm 0.0094}$ & $0.5468_{\pm 0.1377^{\dagger}}$ & $\mathbf{0.0044}_{\pm 0.0084}$ & $\mathbf{0.9951}_{\pm 0.0085}$ & $0.0126_{\pm 0.0066}$ & $0.9064_{\pm 0.0765}$ \\
 &  & 2 & $0.0118_{\pm 0.0109}$ & $0.3498_{\pm 0.0265}$ & $\mathbf{0.0089}_{\pm 0.0034}$ & $0.0286_{\pm 0.0095}$ & $0.0121_{\pm 0.0149}$ & $\mathbf{0.9803}_{\pm 0.0300}$ & $\mathit{0.0077}_{\pm 0.0096}$ & $0.9343_{\pm 0.0690}$ \\
 \cmidrule(lr){2-11}
 & \multirow{3}{*}{SimpleHGN} & 0 & $0.0026_{\pm 0.0051}$ & $0.1320_{\pm 0.0530}$ & $\mathit{0.0013}_{\pm 0.0050}$ & $0.3833_{\pm 0.4501^{\dagger}}$ & $0.0016_{\pm 0.0066}$ & $\mathbf{1.0000}_{\pm 0.0000}$ & $0.0060_{\pm 0.0066}$ & $\mathbf{1.0000}_{\pm 0.0000}$ \\
 &  & 1 & $\mathbf{-0.0026}_{\pm 0.0071}$ & $0.1803_{\pm 0.0829}$ & $0.0039_{\pm 0.0078}$ & $0.6167_{\pm 0.4934^{\dagger}}$ & $\mathit{0.0000}_{\pm 0.0033}$ & $\mathit{0.9787}_{\pm 0.0370}$ & $\mathbf{-0.0077}_{\pm 0.0034}$ & $\mathbf{1.0000}_{\pm 0.0000}$ \\
 &  & 2 & $0.0076_{\pm 0.0131}$ & $0.1419_{\pm 0.0416}$ & $0.0063_{\pm 0.0132}$ & $0.5892_{\pm 0.3604^{\dagger}}$ & $\mathbf{-0.0011}_{\pm 0.0147}$ & $\mathbf{1.0000}_{\pm 0.0000}$ & $\mathbf{-0.0077}_{\pm 0.0100}$ & $\mathbf{1.0000}_{\pm 0.0000}$ \\
\midrule

\multirow{9}{*}{IMDB} & \multirow{3}{*}{HAN} & 0 & $\mathit{0.0262}_{\pm 0.0152}$ & $0.3860_{\pm 0.0802}$ & $0.0309_{\pm 0.0067}$ & $0.0093_{\pm 0.0209}$ & $\mathbf{0.0114}_{\pm 0.0157}$ & $\mathit{0.4268}_{\pm 0.3781^{\dagger}}$ & $0.0400_{\pm 0.0036}$ & $\mathbf{0.9377}_{\pm 0.0500}$ \\
 &  & 1 & $0.0022_{\pm 0.0128}$ & $\mathit{0.4140}_{\pm 0.0627}$ & $0.0144_{\pm 0.0191}$ & $0.0000_{\pm 0.0000}$ & $\mathit{0.0010}_{\pm 0.0050}$ & $0.1947_{\pm 0.2739^{\dagger}}$ & $\mathbf{-0.0026}_{\pm 0.0191}$ & $\mathbf{0.8271}_{\pm 0.0492}$ \\
 &  & 2 & $0.0122_{\pm 0.0146}$ & $0.4318_{\pm 0.0679}$ & $\mathit{-0.0037}_{\pm 0.0098}$ & $0.0972_{\pm 0.1323^{\dagger}}$ & $\mathbf{-0.0135}_{\pm 0.0110}$ & $\mathit{0.6402}_{\pm 0.3129^{\dagger}}$ & $0.0109_{\pm 0.0206}$ & $\mathbf{0.8956}_{\pm 0.0118}$ \\
 \cmidrule(lr){2-11}
 & \multirow{3}{*}{HGT} & 0 & $\mathit{0.0034}_{\pm 0.0159}$ & $0.5617_{\pm 0.0487}$ & $0.0044_{\pm 0.0144}$ & $\mathit{0.9888}_{\pm 0.0122}$ & $0.0057_{\pm 0.0133}$ & $0.8583_{\pm 0.0928}$ & $\mathbf{-0.0042}_{\pm 0.0216}$ & $\mathbf{1.0000}_{\pm 0.0000}$ \\
 &  & 1 & $0.0069_{\pm 0.0226}$ & $0.5953_{\pm 0.0775}$ & $\mathbf{-0.0050}_{\pm 0.0186}$ & $\mathit{0.9346}_{\pm 0.0509}$ & $0.0213_{\pm 0.0162}$ & $0.9159_{\pm 0.0799}$ & $\mathit{0.0000}_{\pm 0.0062}$ & $\mathbf{1.0000}_{\pm 0.0000}$ \\
 &  & 2 & $0.0250_{\pm 0.0277}$ & $0.4654_{\pm 0.0733}$ & $\mathbf{-0.0072}_{\pm 0.0137}$ & $\mathit{0.8280}_{\pm 0.0782}$ & $0.0088_{\pm 0.0234}$ & $0.7461_{\pm 0.1238^{\dagger}}$ & $\mathit{-0.0047}_{\pm 0.0078}$ & $\mathbf{1.0000}_{\pm 0.0000}$ \\
 \cmidrule(lr){2-11}
 & \multirow{3}{*}{SimpleHGN} & 0 & $\mathit{-0.0072}_{\pm 0.0222}$ & $0.7495_{\pm 0.0357}$ & $\mathit{-0.0072}_{\pm 0.0232}$ & $\mathit{0.9551}_{\pm 0.0226}$ & $-0.0026_{\pm 0.0164}$ & $0.9408_{\pm 0.0177}$ & $\mathbf{-0.1477}_{\pm 0.1436^{\dagger}}$ & $\mathbf{1.0000}_{\pm 0.0000}$ \\
 &  & 1 & $\mathit{0.0041}_{\pm 0.0139}$ & $0.8869_{\pm 0.0220}$ & $0.0059_{\pm 0.0202}$ & $\mathit{0.9804}_{\pm 0.0361}$ & $0.0068_{\pm 0.0160}$ & $0.8427_{\pm 0.1139^{\dagger}}$ & $\mathbf{-0.2470}_{\pm 0.0390}$ & $\mathbf{1.0000}_{\pm 0.0000}$ \\
 &  & 2 & $0.0100_{\pm 0.0155}$ & $0.6953_{\pm 0.0534}$ & $\mathit{-0.0094}_{\pm 0.0086}$ & $\mathit{0.9364}_{\pm 0.0403}$ & $0.0192_{\pm 0.0104}$ & $0.7664_{\pm 0.0875}$ & $\mathbf{-0.1607}_{\pm 0.1271^{\dagger}}$ & $\mathbf{1.0000}_{\pm 0.0000}$ \\
\bottomrule
\end{tabular}%
}
\end{table*}

\subsubsection{Train settings}
We conduct experiments using HAN \citep{HAN}, HGT \citep{HGT}, and SimpleHGN \citep{SimpleHGN} as victim models, ensuring a fair comparison of backdoor attack performance under the same training and evaluation conditions. The test accuracy of these victim models on different datasets is reported in Table \ref{tab:clean_accuracy}. The dataset is divided into training, testing, and validation sets. The training set comprises 70\% of the primary-type nodes $\mathcal{V}_{t_p}$, including both clean and poisoned nodes. Specifically, the poisoned training set (Poison Trainset) accounts for 5\% of $\mathcal{V}_{t_p}$, serving as the injected trigger nodes to facilitate backdoor activation. The testing set constitutes 20\% of $\mathcal{V}_{t_p}$, within which the poisoned testing set (Poison Testset) also accounts for 5\%, allowing us to evaluate the attack's effectiveness during inference. The remaining 10\% is allocated to the validation set, which is used for hyperparameter tuning and early stopping. The training parameters are provided in Appendix \ref{appendix_sec:other training parameters}.

\begin{table}[htbp]
\centering
\small
\caption{Test accuracy of victim models on ACM, DBLP, and IMDB datasets.}
\label{tab:clean_accuracy}
\begin{tabular}{lccc}
\toprule
Model & ACM & DBLP & IMDB \\
\midrule
HAN       & 0.8571 & 0.9199 & 0.7263 \\
HGT       & 0.918  & 0.9309 & 0.7380 \\
SimpleHGN & 0.9019 & 0.9396 & 0.7910  \\
\bottomrule
\end{tabular}
\end{table}

\subsubsection{Compared Methods}
To comprehensively evaluate our proposed method, we select distinct baselines tailored for specific evaluation objectives. For performance under defense mechanisms, we employ UGBA \citep{UGBA} and its variant DPGBA \citep{DPGBA}, which utilizes an out-of-distribution detector to ensure trigger distribution consistency. Regarding standard attack effectiveness, we compare our approach against CGBA \citep{CGBA} and HGBA \citep{HBA}. CGBA relies on feature perturbations without structural modifications, while HGBA injects backdoor patterns via specific edges within meta-paths. These baselines provide a rigorous framework for assessing both the robustness and the potency of our proposed attack.

To ensure a fair comparison among these diverse methodologies, we define the attack budget as the total number of poisoned nodes instead of the structural perturbation budget used in some recent studies. This choice prevents the evaluation from favoring lightweight trigger designs that might compromise more nodes under a fixed structural limit. Furthermore, the number of compromised accounts represents the primary resource bottleneck in practice \cite{metric_compare}, and excessively high poisoning ratios are easily detected via statistical anomalies like label distribution shifts. Finally, we adapt the homogeneous baselines to heterogeneous graph structures to maintain experimental consistency.

\subsubsection{Evaluation Metrics}

The \textit{Attack Success Rate (ASR)} \citep{UGBA} measures the probability that the backdoored model \( f_b \) misclassifies a sample embedded with a trigger \( g_t \) into the target class \( y_t \). Formally, ASR is defined as:

\begin{equation}
ASR = \frac{\sum_{i=1}^{n} \mathbf{1}(f_b(v_i) = y_t)}{n}
\end{equation}
where \( n \) denotes the number of poisoned test samples, and \( \mathbf{1}(\cdot) \) represents the indicator function. A higher ASR indicates a more effective backdoor attack.

The \textit{Clean Accuracy Drop (CAD)} quantifies the performance degradation of the backdoored model on benign tasks compared to a clean model. It measures whether the backdoor injection impairs the model's utility on normal inputs. Formally, CAD is defined as:\begin{equation}CAD = Acc_{f_c}(\text{Clean}) - Acc_{f_b}(\text{Clean})\end{equation}where $ Acc_{f_c}(\text{Clean}) $ denotes the classification accuracy of the model trained on clean data (reference model), and $ Acc_{f_b}(\text{Clean}) $ denotes the accuracy of the backdoored model, both evaluated on the clean test set. A lower CAD (ideally close to 0) indicates higher stealthiness, as it suggests the backdoored model maintains performance comparable to the clean baseline on benign samples.

The \textit{Diversity Score} quantifies the variability in the connection patterns of generated trigger nodes. Since trigger nodes connect to multiple types of auxiliary nodes, the final score is the average diversity across all auxiliary relation types. It is defined as:
\begin{equation}
\text{Diversity Score} = \frac{1}{|\mathcal{T}_{aux}|} \sum_{t \in \mathcal{T}_{aux}} \underset{i < j}{\mathbb{E}} \left[ 1 - \frac{p_{i,t}^\top p_{j,t}}{\|p_{i,t}\|_2 \|p_{j,t}\|_2} \right]
\end{equation}
where $ \mathcal{T}_{aux} $ is the set of auxiliary relation types. $ p_{i,t} \in \{0, 1\}^{|C_t|} $ represents the binary connection vector of the $ i $-th trigger for auxiliary type $ t $ (where 1 indicates a connection and 0 otherwise), and $ \|\cdot\|_2 $ denotes the $ L_2 $ norm. The term $ \frac{p_{i,t}^\top p_{j,t}}{\|p_{i,t}\|_2 \|p_{j,t}\|_2} $ computes the cosine similarity between two binary vectors. A higher score implies that triggers exhibit more diverse connection patterns.

\subsection{Experiment result}
\begin{table*}[htbp]
\centering
\caption{Backdoor attack performance comparison (CAD and ASR) under defense mechanism. \textbf{Bold} indicates the best performance (lowest CAD, highest ASR), and \textit{italics} indicates the second-best performance.}
\label{tab:attack_under_defen}
\resizebox{\textwidth}{!}{%
\begin{tabular}{ccc cc cc cc cc}
\toprule
\multirow{2}{*}{Dataset} & \multirow{2}{*}{Model} & \multirow{2}{*}{Class} & \multicolumn{2}{c}{DPGBA} & \multicolumn{2}{c}{UGBA} & \multicolumn{2}{c}{HeteroHBA-Variant $\mathrm{II}$} & \multicolumn{2}{c}{HeteroHBA (Ours)} \\
\cmidrule(lr){4-5} \cmidrule(lr){6-7} \cmidrule(lr){8-9} \cmidrule(lr){10-11}
 &  &  & CAD & ASR & CAD & ASR & CAD & ASR & CAD & ASR \\
\midrule

\multirow{9}{*}{ACM} 
 & \multirow{3}{*}{HAN} 
    & 0 & $-0.0344_{\pm 0.0978}$ & $0.0985_{\pm 0.0400}$ & $\mathbf{-0.1526_{\pm 0.0809}}$ & $0.0537_{\pm 0.0138}$ & $\mathit{-0.0502_{\pm 0.0432}}$ & $\mathbf{0.9867_{\pm 0.0188}}$ & $-0.0568_{\pm 0.0532}$ & $\mathbf{0.9867_{\pm 0.0188}}$ \\
 & & 1 & $\mathit{-0.0437_{\pm 0.1022}}$ & $0.1602_{\pm 0.1181}$ & $\mathbf{-0.0636_{\pm 0.1087}}$ & $0.1502_{\pm 0.1279}$ & $0.0011_{\pm 0.0108}$ & $\mathit{0.9768_{\pm 0.0125}}$ & $\mathit{0.0127_{\pm 0.0063}}$ & $\mathbf{0.9983_{\pm 0.0029}}$ \\
 & & 2 & $\mathbf{-0.0331_{\pm 0.0809}}$ & $0.0199_{\pm 0.0145}$ & $0.0123_{\pm 0.0967}$ & $0.0179_{\pm 0.0120}$ & $0.0204_{\pm 0.0207}$ & $\mathit{0.9884_{\pm 0.0104}}$ & $\mathit{-0.0226_{\pm 0.0163}}$ & $\mathbf{0.9917_{\pm 0.0057}}$ \\
\cmidrule{2-11}
 & \multirow{3}{*}{HGT} 
    & 0 & $\mathbf{-0.0040_{\pm 0.0123}}$ & $0.0627_{\pm 0.0201}$ & $\mathit{0.0033_{\pm 0.0093}}$ & $0.0587_{\pm 0.0215}$ & $0.0055_{\pm 0.0122}$ & $\mathbf{1.0000_{\pm 0.0000}}$ & $\mathit{0.0155_{\pm 0.0069}}$ & $0.9701_{\pm 0.0348}$ \\
 & & 1 & $0.0146_{\pm 0.0201}$ & $0.1313_{\pm 0.0886}$ & $\mathit{0.0063_{\pm 0.0066}}$ & $0.0816_{\pm 0.0255}$ & $0.0039_{\pm 0.0124}$ & $\mathbf{1.0000_{\pm 0.0000}}$ & $\mathbf{0.0017_{\pm 0.0088}}$ & $\mathit{0.9950_{\pm 0.0050}}$ \\
 & & 2 & $\mathbf{-0.0033_{\pm 0.0048}}$ & $0.0050_{\pm 0.0000}$ & $\mathit{0.0077_{\pm 0.0110}}$ & $0.0050_{\pm 0.0035}$ & $0.0099_{\pm 0.0166}$ & $\mathbf{1.0000_{\pm 0.0000}}$ & $0.0079_{\pm 0.0079}$ & $\mathit{0.9900_{\pm 0.0100}}$ \\
\cmidrule{2-11}
 & \multirow{3}{*}{SimpleHGN} 
    & 0 & $0.0026_{\pm 0.0067}$ & $0.0527_{\pm 0.0067}$ & $\mathit{-0.0017_{\pm 0.0065}}$ & $0.0428_{\pm 0.0160}$ & $\mathit{-0.0061_{\pm 0.0051}}$ & $\mathbf{1.0000_{\pm 0.0000}}$ & $\mathbf{-0.0099_{\pm 0.0060}}$ & $0.9751_{\pm 0.0199}$ \\
 & & 1 & $0.0043_{\pm 0.0175}$ & $0.1035_{\pm 0.0221}$ & $0.0096_{\pm 0.0127}$ & $0.1015_{\pm 0.0090}$ & $\mathit{-0.0011_{\pm 0.0110}}$ & $\mathbf{0.9950_{\pm 0.0086}}$ & $\mathbf{-0.0044_{\pm 0.0124}}$ & $\mathit{0.9685_{\pm 0.0546}}$ \\
 & & 2 & $0.0093_{\pm 0.0129}$ & $0.0219_{\pm 0.0109}$ & $0.0132_{\pm 0.0066}$ & $0.0090_{\pm 0.0022}$ & $0.0121_{\pm 0.0117}$ & $\mathbf{1.0000_{\pm 0.0000}}$ & $\mathbf{-0.0094_{\pm 0.0034}}$ & $\mathit{0.9801_{\pm 0.0263}}$ \\
\midrule

\multirow{9}{*}{DBLP} 
 & \multirow{3}{*}{HAN} 
    & 0 & $\mathbf{-0.0043_{\pm 0.0125}}$ & $0.0700_{\pm 0.0118}$ & $\mathit{0.0118_{\pm 0.0073}}$ & $0.0818_{\pm 0.0075}$ & $\mathit{0.0082_{\pm 0.0039}}$ & $0.3103_{\pm 0.3969}$ & $0.0384_{\pm 0.0468}$ & $\mathbf{1.0000_{\pm 0.0000}}$ \\
 & & 1 & $0.0145_{\pm 0.0230}$ & $0.0670_{\pm 0.0202}$ & $\mathbf{0.0016_{\pm 0.0117}}$ & $0.0709_{\pm 0.0102}$ & $\mathbf{0.0016_{\pm 0.0124}}$ & $\mathit{0.1790_{\pm 0.2334}}$ & $\mathit{0.0186_{\pm 0.0038}}$ & $\mathbf{1.0000_{\pm 0.0000}}$ \\
 & & 2 & $\mathbf{-0.0010_{\pm 0.0113}}$ & $0.0749_{\pm 0.0145}$ & $\mathit{0.0026_{\pm 0.0153}}$ & $0.0611_{\pm 0.0190}$ & $\mathit{0.0066_{\pm 0.0202}}$ & $0.3711_{\pm 0.5453}$ & $0.0049_{\pm 0.0059}$ & $\mathbf{1.0000_{\pm 0.0000}}$ \\
\cmidrule{2-11}
 & \multirow{3}{*}{HGT} 
    & 0 & $0.0132_{\pm 0.0164}$ & $0.4424_{\pm 0.2695}$ & $0.0214_{\pm 0.0092}$ & $\mathit{0.3872_{\pm 0.1941}}$ & $\mathbf{0.0041_{\pm 0.0058}}$ & $\mathit{0.2340_{\pm 0.3030}}$ & $0.0077_{\pm 0.0019}$ & $\mathbf{0.9885_{\pm 0.0124}}$ \\
 & & 1 & $0.0145_{\pm 0.0169}$ & $0.1271_{\pm 0.1000}$ & $\mathit{0.0141_{\pm 0.0108}}$ & $\mathit{0.4236_{\pm 0.1360}}$ & $\mathbf{0.0060_{\pm 0.0038}}$ & $0.1363_{\pm 0.1679}$ & $0.0126_{\pm 0.0066}$ & $\mathbf{0.9064_{\pm 0.0765}}$ \\
 & & 2 & $\mathit{0.0043_{\pm 0.0107}}$ & $0.6079_{\pm 0.0243}$ & $0.0092_{\pm 0.0102}$ & $\mathit{0.4394_{\pm 0.1725}}$ & $\mathbf{-0.0022_{\pm 0.0094}}$ & $0.1379_{\pm 0.1621}$ & $\mathit{0.0077_{\pm 0.0096}}$ & $\mathbf{0.9343_{\pm 0.0690}}$ \\
\cmidrule{2-11}
 & \multirow{3}{*}{SimpleHGN} 
    & 0 & $\mathbf{-0.0026_{\pm 0.0055}}$ & $0.3961_{\pm 0.3302}$ & $\mathit{0.0007_{\pm 0.0075}}$ & $\mathit{0.6266_{\pm 0.0273}}$ & $0.0062_{\pm 0.0049}$ & $\mathit{0.4175_{\pm 0.4142}}$ & $0.0060_{\pm 0.0066}$ & $\mathbf{1.0000_{\pm 0.0000}}$ \\
 & & 1 & $0.0026_{\pm 0.0060}$ & $0.5232_{\pm 0.1657}$ & $\mathbf{0.0016_{\pm 0.0088}}$ & $0.5044_{\pm 0.2244}$ & $\mathit{0.0033_{\pm 0.0087}}$ & $\mathit{0.7323_{\pm 0.2425}}$ & $-0.0077_{\pm 0.0034}$ & $\mathbf{1.0000_{\pm 0.0000}}$ \\
 & & 2 & $0.0066_{\pm 0.0103}$ & $0.6394_{\pm 0.0237}$ & $\mathit{-0.0013_{\pm 0.0066}}$ & $0.4030_{\pm 0.2807}$ & $\mathit{-0.0049_{\pm 0.0115}}$ & $\mathit{0.4663_{\pm 0.4959}}$ & $\mathbf{-0.0077_{\pm 0.0100}}$ & $\mathbf{1.0000_{\pm 0.0000}}$ \\
\midrule

\multirow{9}{*}{IMDB} 
 & \multirow{3}{*}{HAN} 
    & 0 & $0.0262_{\pm 0.0152}$ & $\mathit{0.0850_{\pm 0.0061}}$ & $\mathbf{0.0087_{\pm 0.0251}}$ & $\mathit{0.0841_{\pm 0.0074}}$ & $0.0088_{\pm 0.0119}$ & $0.2601_{\pm 0.3050}$ & $\mathit{0.0182_{\pm 0.0211}}$ & $\mathbf{0.9190_{\pm 0.0554}}$ \\
 & & 1 & $\mathit{-0.0037_{\pm 0.0146}}$ & $0.1841_{\pm 0.0205}$ & $\mathbf{-0.0084_{\pm 0.0185}}$ & $\mathit{0.1822_{\pm 0.0119}}$ & $0.0057_{\pm 0.0127}$ & $\mathit{0.1791_{\pm 0.0211}}$ & $0.0026_{\pm 0.0206}$ & $\mathbf{0.8287_{\pm 0.0097}}$ \\
 & & 2 & $\mathit{-0.0012_{\pm 0.0060}}$ & $\mathit{0.1589_{\pm 0.0282}}$ & $\mathbf{-0.0081_{\pm 0.0285}}$ & $0.1449_{\pm 0.0151}$ & $0.0078_{\pm 0.0413}$ & $\mathit{0.1636_{\pm 0.0446}}$ & $0.0166_{\pm 0.0521}$ & $\mathbf{0.8769_{\pm 0.0857}}$ \\
\cmidrule{2-11}
 & \multirow{3}{*}{HGT} 
    & 0 & $\mathit{-0.0047_{\pm 0.0191}}$ & $0.0850_{\pm 0.0167}$ & $\mathbf{-0.0081_{\pm 0.0141}}$ & $\mathit{0.1000_{\pm 0.0445}}$ & $0.0094_{\pm 0.0338}$ & $\mathit{0.1963_{\pm 0.1698}}$ & $0.0010_{\pm 0.0310}$ & $\mathbf{0.8302_{\pm 0.0480}}$ \\
 & & 1 & $\mathbf{-0.0062_{\pm 0.0183}}$ & $0.1813_{\pm 0.0524}$ & $0.0156_{\pm 0.0154}$ & $\mathit{0.1953_{\pm 0.0182}}$ & $0.0161_{\pm 0.0362}$ & $0.1589_{\pm 0.0214}$ & $\mathit{0.0078_{\pm 0.0258}}$ & $\mathbf{0.8723_{\pm 0.0823}}$ \\
 & & 2 & $\mathbf{0.0034_{\pm 0.0195}}$ & $\mathit{0.1729_{\pm 0.0674}}$ & $\mathit{0.0094_{\pm 0.0124}}$ & $0.1766_{\pm 0.0462}$ & $0.0187_{\pm 0.0231}$ & $0.1636_{\pm 0.0243}$ & $0.0343_{\pm 0.0133}$ & $\mathbf{0.8131_{\pm 0.0382}}$ \\
\cmidrule{2-11}
 & \multirow{3}{*}{SimpleHGN} 
    & 0 & $\mathbf{-0.0147_{\pm 0.0049}}$ & $\mathit{0.0757_{\pm 0.0077}}$ & $\mathit{0.0016_{\pm 0.0162}}$ & $\mathit{0.0729_{\pm 0.0117}}$ & $0.0021_{\pm 0.0152}$ & $0.2679_{\pm 0.3106}$ & $0.0182_{\pm 0.0126}$ & $\mathbf{0.8505_{\pm 0.0701}}$ \\
 & & 1 & $\mathbf{-0.0000_{\pm 0.0093}}$ & $0.1336_{\pm 0.0223}$ & $\mathit{0.0059_{\pm 0.0088}}$ & $\mathit{0.1402_{\pm 0.0201}}$ & $0.0099_{\pm 0.0213}$ & $\mathit{0.1417_{\pm 0.0135}}$ & $0.0101_{\pm 0.0216}$ & $\mathbf{0.9735_{\pm 0.0216}}$ \\
 & & 2 & $\mathbf{0.0022_{\pm 0.0134}}$ & $0.1822_{\pm 0.0437}$ & $\mathit{0.0084_{\pm 0.0150}}$ & $\mathit{0.1626_{\pm 0.0434}}$ & $0.0120_{\pm 0.0181}$ & $0.1340_{\pm 0.0177}$ & $0.0157_{\pm 0.0332}$ & $\mathbf{0.9361_{\pm 0.0332}}$ \\
\bottomrule
\end{tabular}
} 
\end{table*}

\subsubsection{Attack effectiveness}
To evaluate the effectiveness of the proposed framework, we compare HeteroHBA against baselines CGBA and HGBA, along with an ablation variant, HeteroHBA-Variant I. Inspired by HGBA, this variant bypasses trigger injection, directly selecting existing neighbors via similar candidate pool and bi-level optimization mechanisms. Table~\ref{tab:attack_performance} summarizes the results, leading to the following observations.

First, HeteroHBA outperforms both CGBA and HGBA in Attack Success Rate (ASR) across most scenarios. Specifically, CGBA yields relatively low scores, indicating the ineffectiveness of feature-based attacks in heterogeneous graphs. HGBA, while competitive in specific cases, suffers from significant instability. As shown in Table~\ref{tab:attack_performance}, it frequently exhibits high standard deviations, suggesting a lack of robustness against initialization randomness or specific topologies. Second, HeteroHBA demonstrates superior robustness compared to HeteroHBA-Variant I. Relying on existing neighbors with uncontrollable features, the variant exhibits high variance and suboptimal ASR in certain scenarios, despite being competitive elsewhere. This underscores the critical role of the injected trigger node as a stable semantic mediator. Finally, regarding Clean Accuracy Drop (CAD), HeteroHBA consistently ranks as the second-best solution with negligible gaps compared to the top performer. This confirms that HeteroHBA achieves a desirable trade-off, delivering high attack potency with minimal utility degradation.
  \begin{figure*}[htbp]
    \centering
    \begin{subfigure}[b]{0.24\textwidth}
      \centering
      \includegraphics[width=\linewidth]{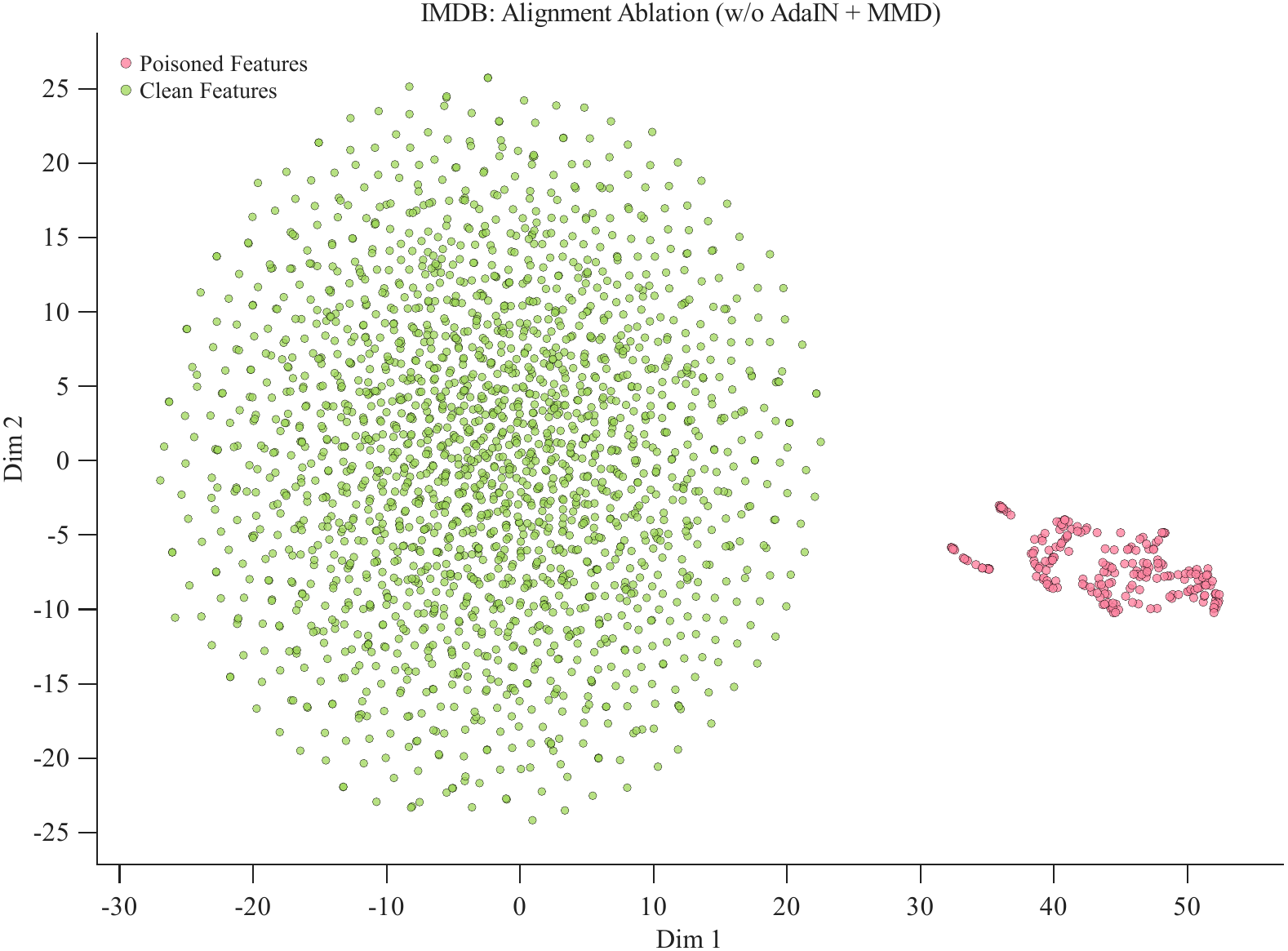}
    \end{subfigure}
    \hfill
    \begin{subfigure}[b]{0.24\textwidth}
      \centering
      \includegraphics[width=\linewidth]{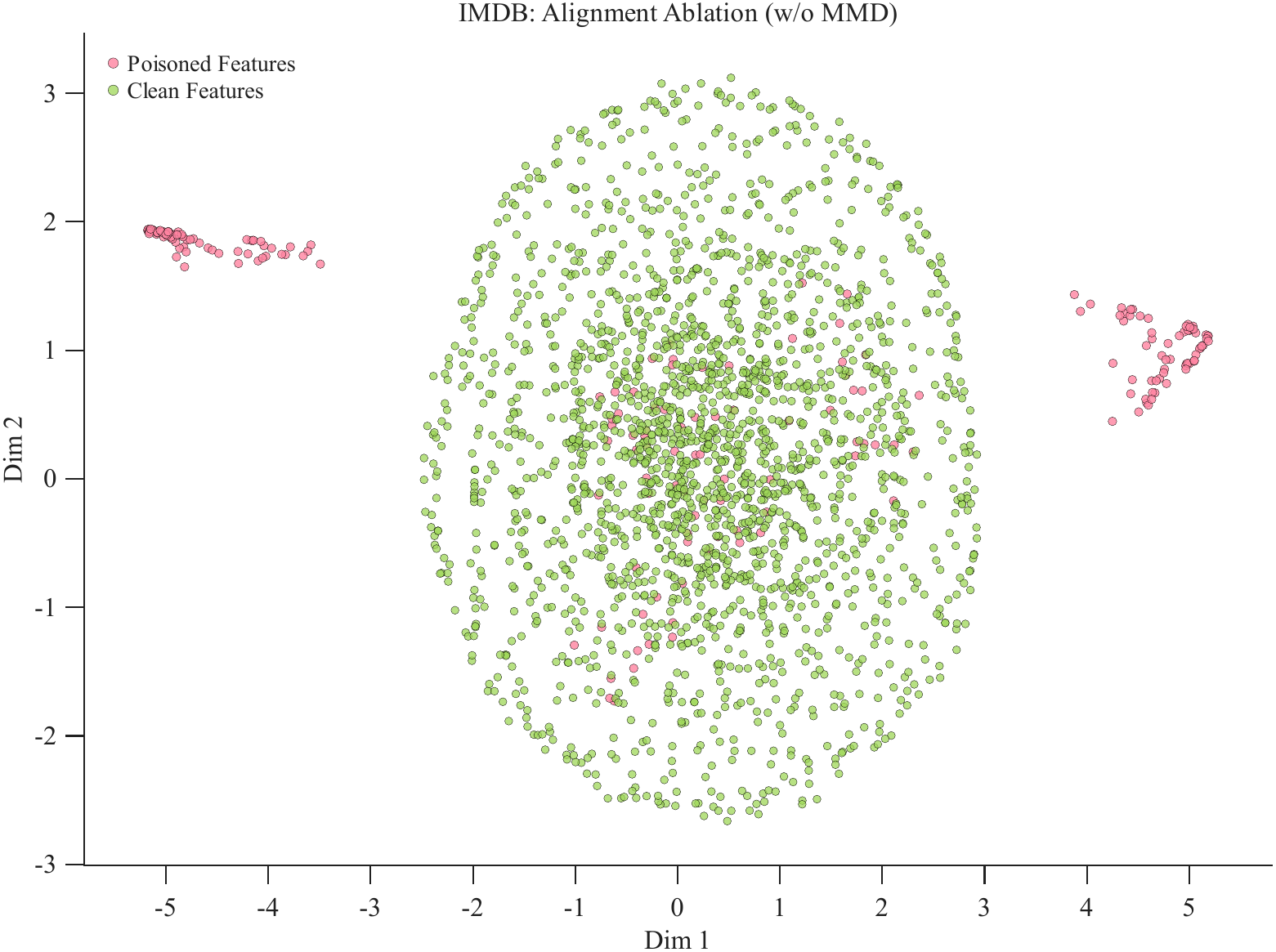}
    \end{subfigure}
    \hfill
    \begin{subfigure}[b]{0.24\textwidth}
      \centering
      \includegraphics[width=\linewidth]{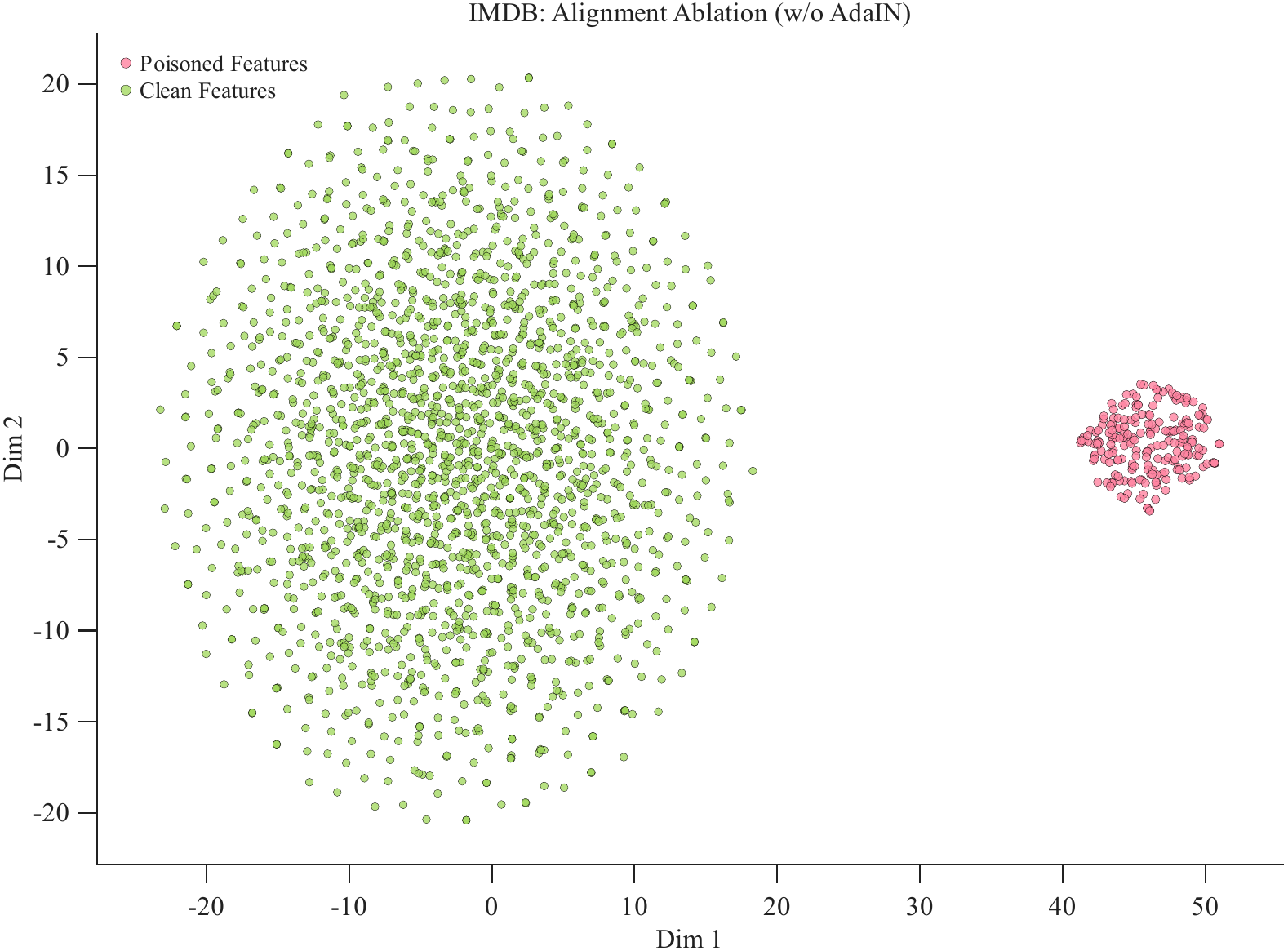}
    \end{subfigure}
    \hfill
    \begin{subfigure}[b]{0.24\textwidth}
      \centering
      \includegraphics[width=\linewidth]{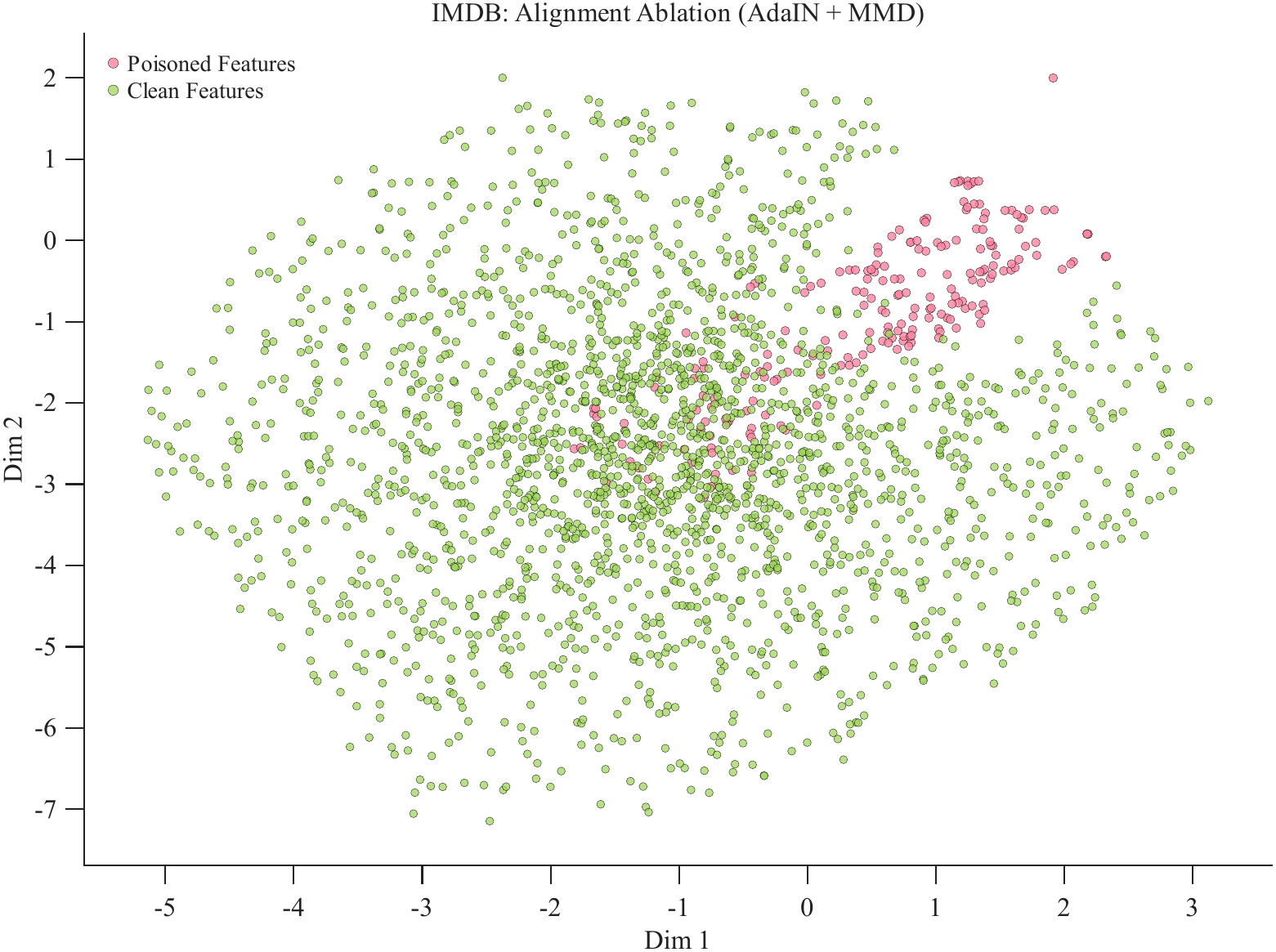}
    \end{subfigure}
    \caption{Alignment ablation results on IMDB. Left-to-right: w/o AdaIN+MMD, w/o MMD, w/o AdaIN, AdaIN+MMD.}
    \label{fig:ablation_tsne}
  \end{figure*}
\subsubsection{Defense Resistance Analysis (RQ2)}
To evaluate the robustness of our proposed attack against potential countermeasures, we introduce Cluster-based Structural Defense (CSD), a defense strategy tailored for heterogeneous graphs. CSD is motivated by the fundamental limitations of existing defenses designed for homogeneous settings. Most graph pruning methods rely on the homophily assumption and use cosine similarity between node features to identify suspicious edges \cite{UGBA, DPGBA}, which often breaks down in heterogeneous graphs where connected nodes belong to different types (e.g., Author and Paper) and are represented in distinct feature spaces. In such cases, benign heterogeneous neighbors may exhibit inherently low cosine similarity, while high cosine similarity alone does not guarantee Euclidean closeness, causing similarity-based pruning to both erroneously remove valid edges and miss poisoned connections. OD-based defenses suffer from complementary limitations: their effectiveness is highly sensitive to threshold selection, which implicitly assumes knowledge of the contamination level (e.g., pruning the top-$k\%$ samples) or access to clean supervision. In blind defense settings, heuristic thresholds often lead to false positives or missed detections. Moreover, these methods require training an autoencoder, introducing non-trivial computational overhead, and focus solely on anomaly detection without correcting corrupted node labels, resulting in irreversible information loss that further degrades downstream learning performance.

To address these challenges, CSD operates in a strictly unsupervised manner by leveraging intrinsic feature distributions of heterogeneous nodes. We observe that nodes injected by homogeneous-based backdoor attacks often deviate from benign nodes of the same type in feature space. Accordingly, CSD is designed as a two-stage process:

First, we project the high-dimensional features of each node type into a latent space using Principal Component Analysis (PCA) and apply a 2-Means clustering algorithm. We quantify the separation between the two resulting clusters using a separation ratio $R$ \cite{csd}, defined as:
\begin{equation}
    R = \frac{\| \boldsymbol{\mu}_1 - \boldsymbol{\mu}_2 \|_2}{\sigma_1 + \sigma_2},
\end{equation}
where $\boldsymbol{\mu}_i$ and $\sigma_i$ denote the centroid and the root-mean-square (RMS) radius of cluster $i$, respectively. Following the empirical guidelines in \cite{csd}, we set the predefined threshold to 2. If $R$ exceeds a predefined threshold, indicating a significant distributional divergence, the smaller cluster is identified as the set of suspicious trigger nodes. CSD then prunes all edges connected to these suspicious nodes to sever the propagation path of the backdoor pattern.

 To mitigate the impact of potentially corrupted labels on victim nodes, CSD incorporates a label rectification mechanism. For any suspicious victim node (i.e., a target node previously connected to a pruned trigger), we perform a $k$-Nearest Neighbors ($k$-NN) search within the set of same-type nodes based on feature similarity. The label of the victim node is then updated via majority voting from its clean neighbors. This ensures that both the graph structure and the supervisory signals are purified without requiring external clean data.

Table \ref{tab:attack_under_defen} reports the performance of different attack methods under the Cluster-based Structural Defense (CSD). In this experiment, we adapt two homogeneous baselines (DPGBA and UGBA) and introduce a HeteroHBA variant (Variant II), which employs an autoencoder to keep the generated trigger nodes in-distribution instead of using AdaIN and MMD loss. The results show that DPGBA and UGBA become largely ineffective under CSD, while HeteroHBA Variant II, despite achieving promising performance on the ACM dataset, fails to generalize to the other two datasets. In contrast, the proposed HeteroHBA demonstrates consistent robustness across all evaluation scenarios. Furthermore, for the pruning-based defense, we derive the cosine similarity distribution over all edges and remove a small fraction of the least similar ones, while for the OD-based defense, we train a separate autoencoder for each node type and discard a small fraction of nodes with the highest reconstruction loss. As shown in Table \ref{tab:other_defense_comparison}, in most cases the impact on both ASR and CAD remains minor, indicating that our attack is largely unaffected by these defense strategies.

\begin{table}[htbp]
\centering
\small
\setlength{\tabcolsep}{4pt}
\caption{Performance comparison under other defense mechanisms.}
\label{tab:other_defense_comparison}
\begin{tabular}{cccccc}
\toprule
Dataset & Model & \multicolumn{2}{c}{OD} & \multicolumn{2}{c}{Prune} \\
\cmidrule(lr){3-4} \cmidrule(lr){5-6}
 &  & CAD & ASR & CAD & ASR \\
\midrule
\multirow{3}{*}{ACM} 
& HAN       & -0.0120 & 0.8843 & -0.0145 & 0.8968 \\
& HGT       & -0.0036 & 0.8776 & 0.0099  & 0.8905 \\
& SimpleHGN & -0.0225 & 0.8955 & 0.0020  & 0.8846 \\
\midrule
\multirow{3}{*}{DBLP} 
& HAN       & 0.0168 & 0.9970 & 0.0194 & 0.9635 \\
& HGT       & 0.0143 & 0.9951 & 0.0059 & 0.9281 \\
& SimpleHGN & 0.0102 & 0.9793 & 0.0063 & 0.9616 \\
\midrule
\multirow{3}{*}{IMDB} 
& HAN       & 0.0224 & 0.6791 & 0.0381 & 0.8981 \\
& HGT       & -0.1061 & 0.7227 & 0.0175 & 0.7336 \\
& SimpleHGN & 0.0099 & 0.7305 & 0.0090 & 0.8093 \\
\bottomrule
\end{tabular}
\end{table}

\begin{table*}[htbp]
\centering
\small
\setlength{\tabcolsep}{6pt}
\renewcommand{\arraystretch}{1.2}
\caption{Mean CAD / ASR under different candidate pool sizes.}
\begin{tabular}{lccccccc}
\hline
\makecell[l]{Dataset /\\ Candidate Pool Fold} & 2 & 4 & 6 & 8 & 10 & 12 & 16 \\
\hline
ACM  & 0.0028 / 0.9884 & 0.0039 / 0.9967 & 0.0000 / 1.0000 & 0.0166 / 1.0000 & 0.0099 / 1.0000 & 0.0022 / 1.0000 & 0.0028 / 0.9967 \\
DBLP & 0.0022 / 1.0000 & 0.0027 / 0.9737 & 0.0027 / 0.9934 & 0.0066 / 0.9975 & 0.0049 / 1.0000 & 0.0107 / 1.0000 & -0.0008 / 1.0000 \\
IMDB & -0.1316 / 1.0000 & -0.1602 / 1.0000 & -0.1747 / 1.0000 & -0.2272 / 1.0000 & -0.2444 / 1.0000 & -0.1019 / 1.0000 & -0.0655 / 1.0000 \\
\hline
\end{tabular}
\label{tab:cad_asr_candidate_pool_size}
\end{table*}

\subsubsection{Hyperparameter Analysis (RQ3)}
\begin{figure}[htbp]
    \centering
    \begin{subfigure}{0.48\linewidth}
        \centering
      \includegraphics[width=\linewidth]{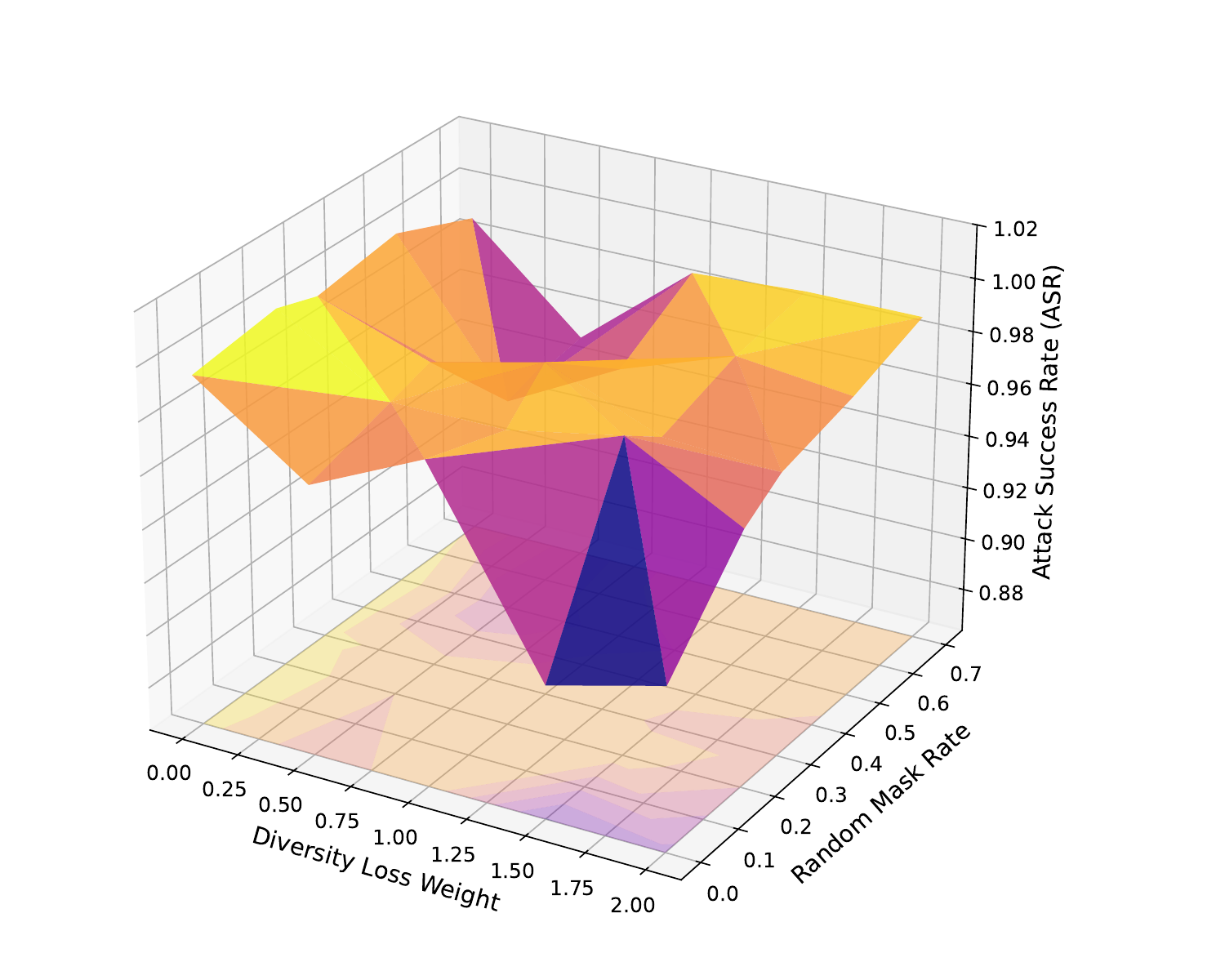}
        \caption{Impact of Hyperparameters on ASR (ACM)}
    \end{subfigure}
    \hfill
    \begin{subfigure}{0.48\linewidth}
        \centering
      \includegraphics[width=\linewidth]{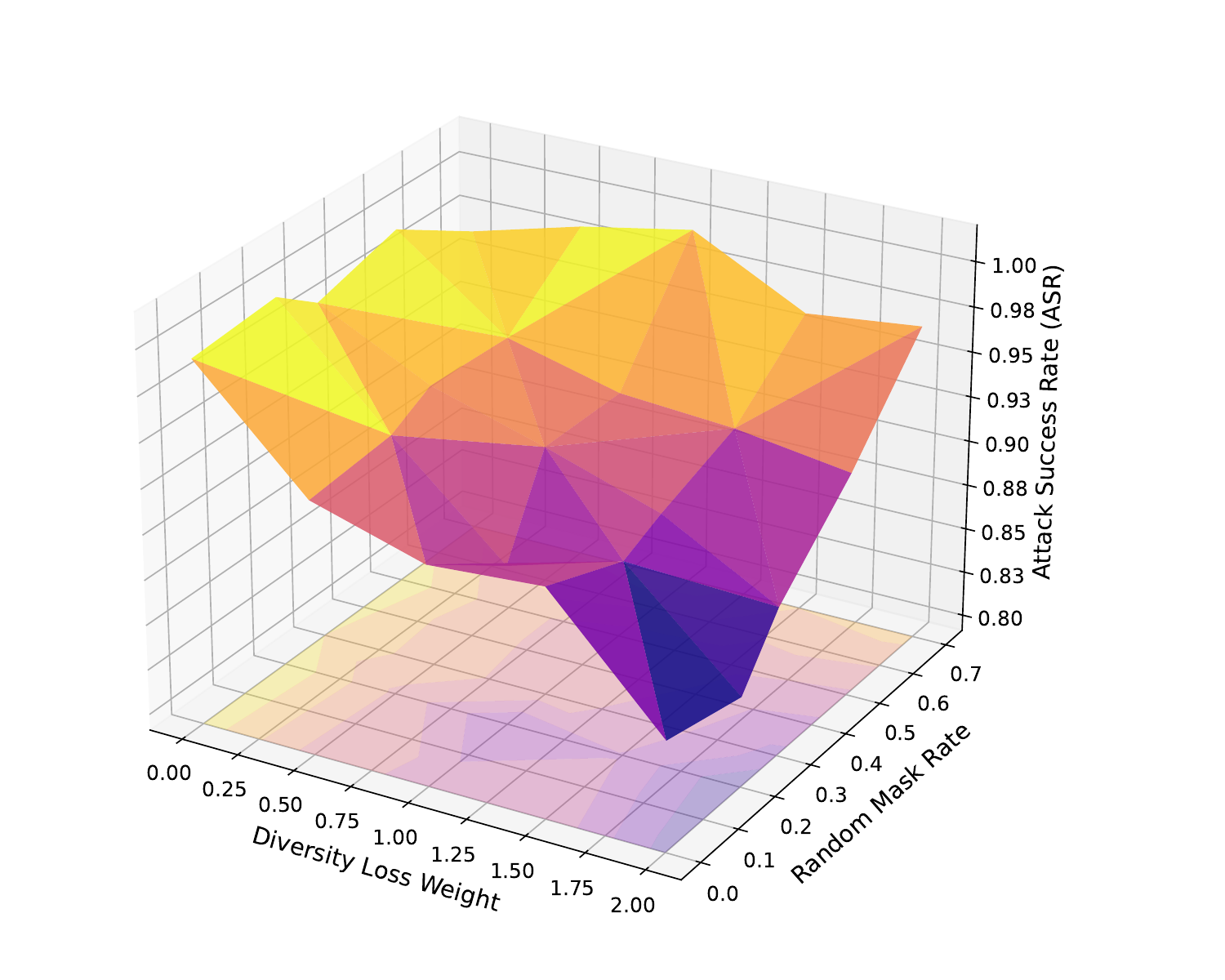}
        \caption{Impact of Hyperparameters on ASR (IMDB)}
    \end{subfigure}
    \caption{Hyperparameter Sensitivity Analysis: Attack Success Rate.}
    \label{fig:hyperparameter_analysis_asr}
\end{figure}

In this section, we analyze the sensitivity of the Random Mask Rate and Diversity Loss Weight on the Attack Success Rate (ASR). As shown in Figure \ref{fig:hyperparameter_analysis_asr} (3D surface plots on IMDB and ACM), our method is largely robust to variations in these hyperparameters: across most of the parameter space, ASR remains consistently high (generally above 90\%), and on ACM it stays close to 100\% over a broad range of configurations. This indicates that the attack effectiveness is not heavily dependent on precise parameter tuning. We nevertheless observe a localized valley where ASR drops more noticeably when the Diversity Loss Weight is set to a mid-to-high value while the Random Mask Rate is low-to-moderate. In this region, the ASR reaches its minimum at approximately 0.80 on IMDB, whereas the lowest ASR on ACM remains higher (around 0.88–0.90), indicating a milder degradation. Overall, these results suggest only limited coupling effects between the two hyperparameters in a narrow region, while stable and strong attack performance is maintained in general.

We further analyze the effect of the candidate pool fold on CAD and ASR (Table \ref{tab:cad_asr_candidate_pool_size}). Overall, varying the fold from 2 to 16 leads to only minor fluctuations: ASR remains close to 1.0 across datasets, and CAD changes only slightly, indicating a limited impact of this setting on the overall performance.

\begin{figure}[htbp]
    \centering
    \begin{subfigure}{0.48\linewidth}
        \centering
        \includegraphics[width=\linewidth]{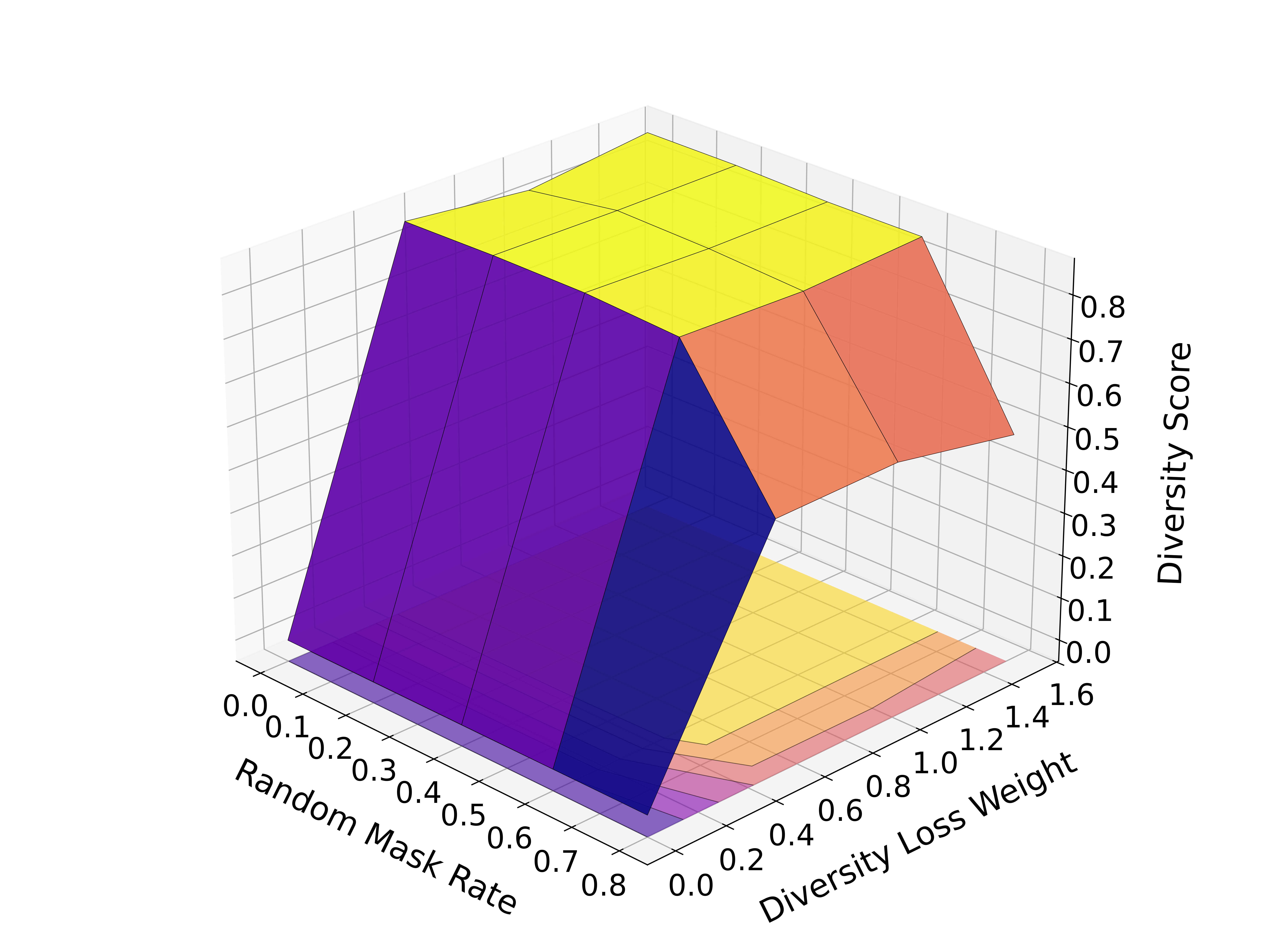}
        \caption{Impact of Hyperparameters on Diversity Score (ACM).}
        \label{fig:hyper_m}
    \end{subfigure}
    \hfill
    \begin{subfigure}{0.48\linewidth}
        \centering
        \includegraphics[width=\linewidth]{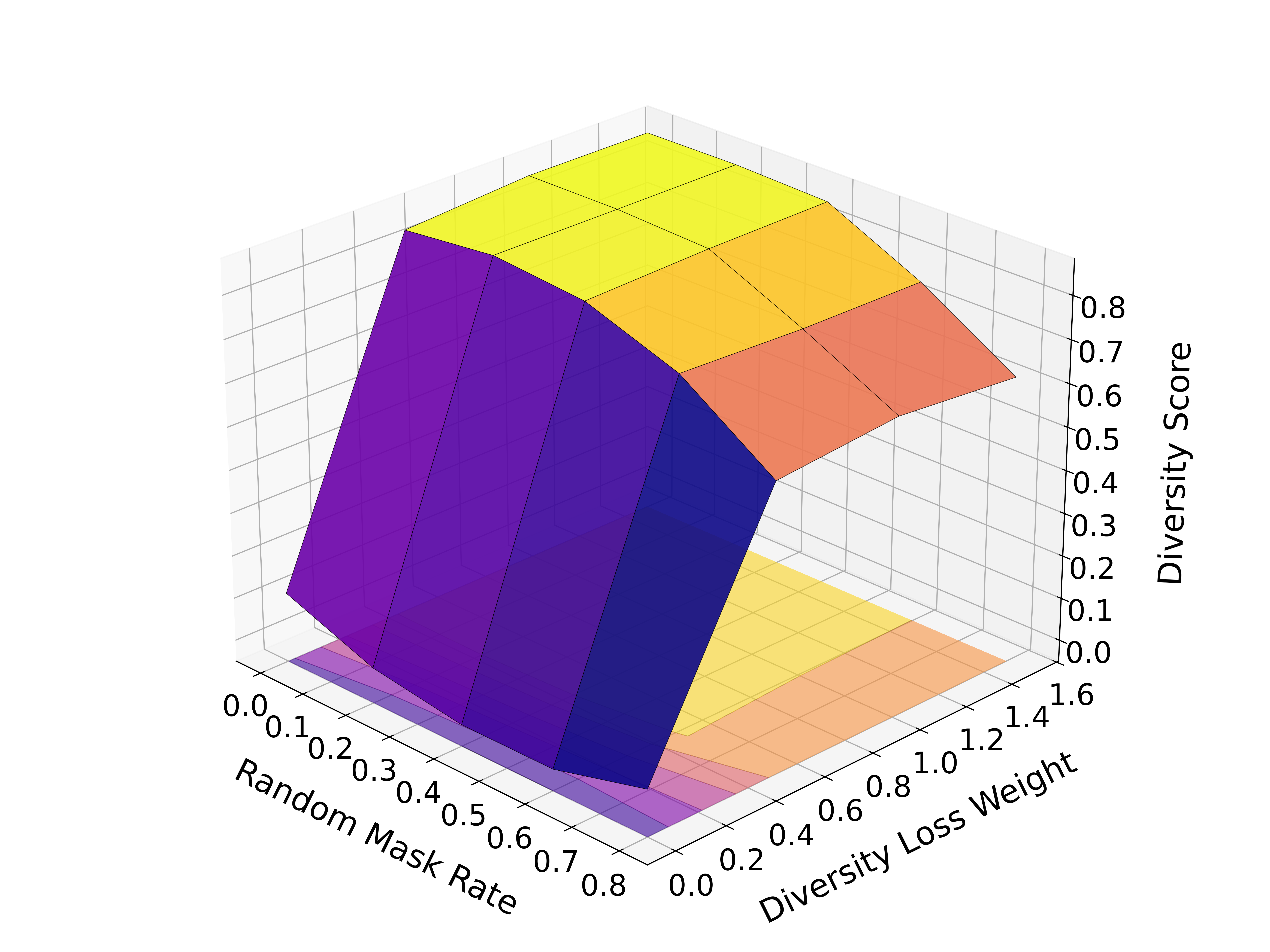}
        \caption{Impact of Hyperparameters on Diversity Score (IMDB).}
        \label{fig:hyper_k}
    \end{subfigure}
    \caption{Hyperparameter Sensitivity Analysis: Diversity Score.}
    \label{fig:hyperparameter_analysis_diversity}
\end{figure}

\subsubsection{Ablation Study}
To address RQ4 and evaluate the individual contributions of HeteroHBA's core components, we conduct some ablation analysis focusing on the feature alignment modules and the saliency-based candidate selection mechanism. This study verifies how each component affects stealthiness and attack performance.

As shown in Fig.~\ref{fig:ablation_tsne}, under the w/o AdaIN+MMD setting, trigger nodes are strongly separated from benign nodes, forming a compact cluster far from the main distribution, which indicates that triggers cannot be well aligned without explicit alignment constraints. Under w/o AdaIN (retaining only MMD), trigger nodes still collapse into an isolated cluster, suggesting that distribution-level matching alone is insufficient. In contrast, under w/o MMD (retaining only AdaIN), while a subset of trigger nodes overlaps with benign ones, noticeable outlier clusters persist. Finally, when both AdaIN and MMD are enabled, trigger nodes are thoroughly mixed with benign nodes. These results demonstrate that AdaIN and MMD play complementary and indispensable roles in achieving seamless feature alignment.

\begin{figure}[htbp] 
  \centering
  \includegraphics[width=\linewidth]{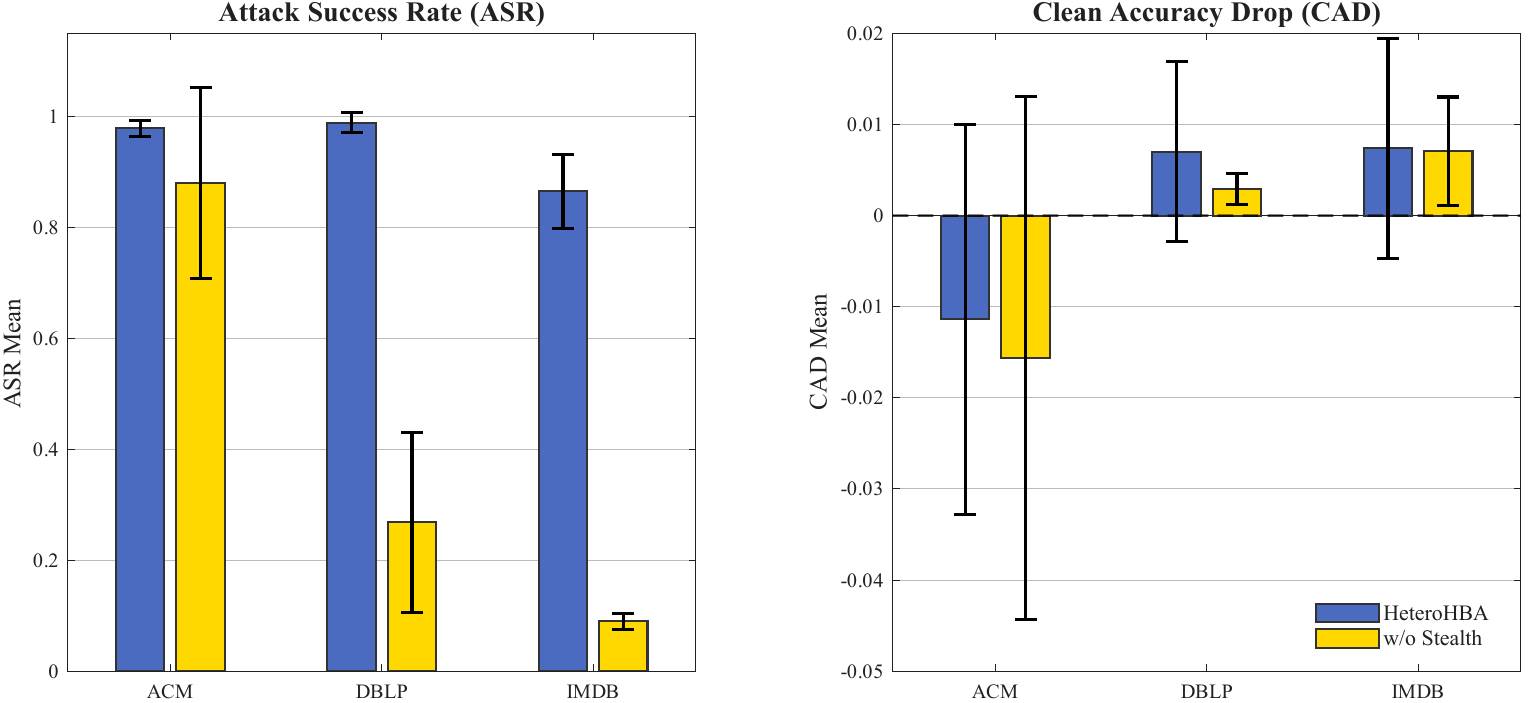}
  \caption{Performance comparison of models with and without the AdaIN and MMD modules.}
  \label{fig:adain_compare}
\end{figure}

The practical impact of this alignment on defense resistance is further quantified in Fig. \ref{fig:adain_compare}. We observe that removing these alignment components causes a sharp decline in the Attack Success Rate (ASR) when facing the Cluster-based Structural Defense (CSD). This drop occurs because, without rigorous statistical constraints, the resulting trigger outliers are easily identified and pruned by heterogeneity-aware defenses. This confirms that feature alignment is not only essential for visual stealthiness but also critical for maintaining attack potency against advanced defense mechanisms.

\begin{figure}[htbp] 
  \centering
  \includegraphics[width=\linewidth]{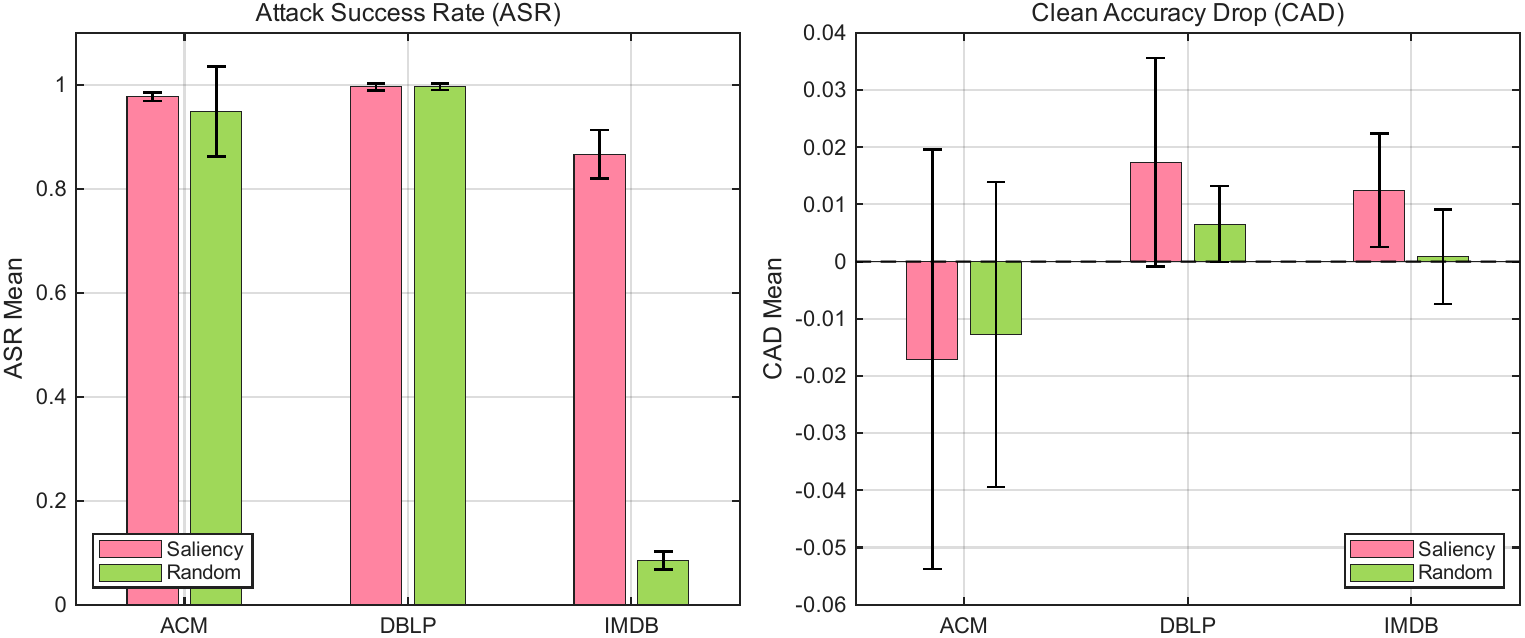}
  \caption{Performance comparison between Saliency and Random selection methods.}
  \label{fig:saliency_vs_random}
\end{figure}

Beyond feature alignment, we evaluate the contribution of the saliency-based candidate selection mechanism. As illustrated in Fig. \ref{fig:saliency_vs_random}, replacing saliency-based screening with random selection leads to varying degrees of ASR reduction across all datasets. This performance degradation is most pronounced in the IMDB dataset, where the ASR drops significantly compared to the full HeteroHBA framework. These findings demonstrate that strategically identifying and connecting to the most influential auxiliary nodes is vital for maximizing adversarial information propagation and ensuring the overall success of the attack.

\section{Conclusion and future work }
In this paper, we proposed HeteroHBA, a novel backdoor attack framework for heterogeneous graphs that utilizes saliency-based candidate selection and a generative GraphTrojanNet with AdaIN and MMD to craft stealthy, effective triggers. Extensive experiments show that HeteroHBA achieves higher attack success rates than state-of-the-art baselines while maintaining high clean accuracy, and remains effective under heterogeneous defenses such as our Cluster-based Structural Defense (CSD). Despite these results, we observe that edge-perturbation-based defenses can noticeably reduce the attack’s effectiveness. Future work will therefore focus on improving attack resilience by developing stronger and more general designs that remain effective under such defenses.
\bibliographystyle{ACM-Reference-Format}
\bibliography{sample-base}

\appendix
\section{HeteroHBA Pseudocode}
\label{appendix_sec: pseudocode}
The pesudocode of HeteroHBA training procedure is provided in Algorithm \ref{alg:heterohba_native}.

\begin{algorithm}[h]
\small
\caption{HeteroHBA Training Procedure (with Post-Generative Refinement)}
\label{alg:heterohba_native}
\begin{algorithmic}[1] 

\REQUIRE Heterogeneous Graph $G$, Target Class $y_t$, Target Nodes $\mathcal{V}^{(p)}$, Params $N, \lambda_{div}$.
\ENSURE Optimized Generator $g_{\theta^*}$, IDA-AT parameters $\{W^*, b^*\}$, Poisoned Graph $\tilde{G}$.

\STATE Identify target-class nodes $\mathcal{V}_{y_t}$ and collect raw pool $\mathcal{C}_{t_a}$ (Eq. \ref{equ:auxiliary_node_calculation}- \ref{equ:raw_candidate_pool_calculation}).
\STATE Calculate pool size $K^{(pool)}_{t_a}$ based on P90 degree (Eq. \ref{equ:candidate_pool_size1}-\ref{equ:candidate_pool_size2}).
\STATE Compute saliency scores $S(v_{t_a})$ and select top-$K$ nodes $\mathcal{C}^*_{t_a}$ (Eq. \ref{equ:saliency_calculation}).

\STATE Initialize generator $\theta_g$ and surrogate $\theta_s$.
\WHILE{not converged}
    \FOR{$n = 1$ to $N$}
        \STATE Generate raw trigger features $\hat{x}_{t_{tr}}^{(i)}$ via AdaIN (Eq. 10).
        \STATE Generate edges $E^{(\mathrm{new})}$ using current $\theta_g$.
        \STATE Update $\theta_s \leftarrow \theta_s - \alpha_s \nabla_{\theta_s} \mathcal{L}_s(\theta_s, \theta_g)$ (Eq. 15).
    \ENDFOR
    
    \STATE Generate triggers on $\mathcal{V}^{(p)}$ using current $\theta_g$.
    \STATE Compute Attack Loss $\mathcal{L}_g$ (Eq. 17) and Diversity Loss $\mathcal{L}_{div}$ (Eq. 14).
    \STATE Update $\theta_g \leftarrow \theta_g - \alpha_g \nabla_{\theta_g} (\mathcal{L}_g + \lambda_{div} \mathcal{L}_{div})$.
\ENDWHILE
\STATE Fix the optimized generator parameters $\theta_g^*$.

\STATE Initialize IDA-AT parameters $W$ and $b$. 
\WHILE{not converged}
  \STATE Apply IDA-AT: $x_{aff} = W \hat{x}_{t_{tr}} + b$. 
    \STATE Compute MMD loss $\mathcal{L}_{MMD}$ for statistical stealthiness (Eq. 11).
    \STATE Compute attack alignment loss $\mathcal{L}_{atk\_aff}$ using surrogate $f_s$ (Eq. 12).
    \STATE Calculate total post-processing loss $\mathcal{L} = \mathcal{L}_{MMD} + \mathcal{L}_{atk\_aff}$.
    \STATE Update $\{W, b\} \leftarrow \{W, b\} - \alpha_{aff} \nabla_{\{W, b\}} \mathcal{L}$.
\ENDWHILE

\RETURN Trained $g_{\theta^*}$ and $\{W^*, b^*\}$.
\end{algorithmic}
\end{algorithm}

\section{CSD Pseudocode}
The pesudocode of HeteroHBA training procedure is provided in Algorithm \ref{alg:csd_defense}.
\begin{algorithm}[h]
\small
\caption{Cluster-based Structural Defense (CSD) Procedure}
\label{alg:csd_defense}
\begin{algorithmic}[1] 

\REQUIRE Poisoned Heterogeneous Graph $\tilde{G}=(\tilde{\mathcal{V}}, \tilde{\mathcal{E}}, \tilde{X})$, Primary Node Type $t_p$, Poisoned Labels $Y_{poison}$, Threshold $\tau$, Neighbor count $k$.
\ENSURE Purified Graph $G^*$, Rectified Labels $Y^*$.

\STATE Initialize victim nodes set $\mathcal{V}_{victim} \leftarrow \emptyset$.
\STATE Initialize purified graph $G^* \leftarrow \tilde{G}$.

\FOR{each node type $t \in \mathcal{T}$}
    \STATE Project features $X_t$ to latent space using PCA and perform 2-Means clustering to obtain clusters $C_1, C_2$.
    \STATE Calculate centroids $\mu_{1,2}$, RMS radii $\sigma_{1,2}$, and Separation Ratio $R$ (Eq. 21).
    
    \IF{$R > \tau$}
        \STATE Identify suspicious trigger cluster $S_{suspicious} \leftarrow \arg\min_{C \in \{C_1, C_2\}} |C|$.
        \STATE Identify primary nodes connected to $S_{suspicious}$ and add to $\mathcal{V}_{victim}$.
        \STATE Prune all edges in $\tilde{\mathcal{E}}$ incident to nodes in $S_{suspicious}$ from $G^*$.
    \ENDIF
\ENDFOR

\FOR{each node $v \in \mathcal{V}_{victim}$}
    \STATE Find $k$-nearest neighbors $\mathcal{N}_k(v)$ in $\mathcal{V}_{t_p}$ based on feature similarity (excluding $S_{suspicious}$).
    \STATE Rectify label $y^*_v \leftarrow \text{MajorityVote}(\{Y_{poison}[u] \mid u \in \mathcal{N}_k(v)\})$.
    \STATE Update label for $v$ in $Y^*$.
\ENDFOR

\RETURN Purified Graph $G^*$, Rectified Labels $Y^*$
\end{algorithmic}
\end{algorithm}

\section{Gradients of the differentiable top-$k$ operator}
\label{appendix:topk_proof}
The following proof is adapted from the blog post by \citep{ahle2022differentiable}.
We consider the differentiable relaxation of the top-$k$ operator defined as follows. 
Let $x = (x_1,\dots,x_n) \in \mathbb{R}^n$ be the input and let 
$\sigma : \mathbb{R} \to (0,1)$ be a smooth and strictly increasing squashing function, 
such as the logistic sigmoid. 
For a fixed integer $k$, define a scalar shift $t = t(x)$ implicitly by the constraint
\begin{equation}
    \sum_{i=1}^n \sigma(x_i + t) = k,
    \label{eq:constraint}
\end{equation}
and define the relaxed top-$k$ output as 
\[
    f_i(x) := \sigma(x_i + t), \qquad i = 1,\dots,n.
\]
Since the left-hand side of \eqref{eq:constraint} is strictly increasing in $t$, 
there exists a unique solution $t(x)$ depending smoothly on~$x$.

To compute the gradient of $f(x)$, differentiate \eqref{eq:constraint} with respect 
to a coordinate~$x_j$. Using the chain rule,
\[
    0
    = \frac{\partial}{\partial x_j} 
      \sum_{i=1}^n \sigma(x_i + t)
    = \sum_{i=1}^n \sigma'(x_i + t) \bigl( \delta_{ij} + \tfrac{\partial t}{\partial x_j} \bigr),
\]
where $\delta_{ij}$ denotes the Kronecker delta. 
Isolating $\partial t / \partial x_j$ yields
\begin{equation}
    \frac{\partial t}{\partial x_j}
    = - \frac{\sigma'(x_j + t)}
             {\sum_{i=1}^n \sigma'(x_i + t)}.
    \label{eq:dt}
\end{equation}

Now differentiate $f_i(x) = \sigma(x_i + t)$ with respect to $x_j$:
\[
    \frac{\partial f_i}{\partial x_j}
    = \sigma'(x_i + t)\bigl(\delta_{ij} + \tfrac{\partial t}{\partial x_j}\bigr).
\]
Substituting \eqref{eq:dt} gives the explicit Jacobian entries
\[
    \frac{\partial f_i}{\partial x_j}
    = \sigma'(x_i + t)\delta_{ij}
      - \sigma'(x_i + t)\,
        \frac{\sigma'(x_j + t)}
             {\sum_{\ell=1}^n \sigma'(x_\ell + t)}.
\]

For compactness, define $v \in \mathbb{R}^n$ by $v_i := \sigma'(x_i + t)$ and 
$\|v\|_1 = \sum_{i=1}^n v_i$. 
Then the Jacobian matrix $J$ with entries $\partial f_i / \partial x_j$ can be written as
\begin{equation}
    J
    = \mathrm{diag}(v)
      - \frac{v v^\top}{\|v\|_1}.
    \label{eq:jacobian}
\end{equation}
This form allows efficient backpropagation: given an upstream gradient vector 
$r \in \mathbb{R}^n$, the vector--Jacobian product is
\[
    r^\top J
    = r^\top \mathrm{diag}(v)
      - \frac{(r^\top v)}{\|v\|_1} v^\top,
\]
which is equivalent to the elementwise expression $r \odot v - \frac{\langle r, v\rangle}{\|v\|_1} v$, 
where $\odot$ denotes the Hadamard product.  This provides a stable and efficient gradient 
for the differentiable top-$k$ transformation.

\begin{table}[t] 
\centering
\caption{Training Parameters and Model-Specific Hyperparameters}
\label{tab:combined_params_simplified}
\resizebox{\linewidth}{!}{
    \begin{tabular}{ll}
    \toprule
    \multicolumn{2}{c}{\textbf{Key Training Parameters}} \\
    \midrule
    \textbf{Parameter}      & \textbf{Value}      \\
    \midrule
    Loss function           & Cross Entropy       \\
    Optimizer               & AdamW               \\
    Epochs                  & 400                 \\
    Learning rate           & 1e-3                \\
    Scheduler               & OneCycleLR          \\
    Dropout                 & 0.2                 \\
    Weight decay            & 1e-4                \\
    Gradient Clipping       & 1.0                 \\
    Candidate Pool Surrogate Model & MAGNN \\ 
    Bi-level Optimization Surrogate Model & HGT \\ 
    \midrule
    \multicolumn{2}{c}{\textbf{Victim Model-Specific Hyperparameters}} \\
    \midrule
    \textbf{Model}  & \textbf{Hyperparameters (Value)} \\
    \midrule
    HGT           & Hidden Units: 64; Layers: 8; Heads: 4 \\
    HAN           & Hidden Units: 64; Layers: 1; Heads: 2 \\
    SimpleHGN     & Hidden Units: 64; Layers: 4; Heads: 8 \\
    \bottomrule
    \end{tabular}
}
\end{table}

\section{Time Complexity Analysis}
\label{appendix_sec:time complexity analysis}

The time complexity of HeteroHBA comprises two main phases: candidate pool construction and bi-level optimization. The construction of the candidate pool proceeds in three sequential steps. First, identifying 2-hop neighbors requires traversing the graph structure, incurring a complexity of $O(|\mathcal{V}_{y_t}| \cdot \bar{D}^2)$, where $|\mathcal{V}_{y_t}|$ is the number of target-class primary nodes and $\bar{D}$ denotes the average degree. Given that real-world heterogeneous graphs typically exhibit long-tailed degree distributions, $\bar{D}$ remains small, ensuring efficiency. Second, the saliency calculation involves training a surrogate model and performing backward passes; we denote this computational cost as $C_{sur}$, which acts as a constant overhead depending on the specific surrogate architecture. Finally, sorting and selecting the top nodes from the raw candidate pool $\mathcal{C}_{t_a}$ takes $O(|\mathcal{C}_{t_a}| \log |\mathcal{C}_{t_a}|)$. Thus, the total complexity for this pre-computation stage is $O(|\mathcal{V}_{y_t}| \cdot \bar{D}^2 + C_{sur} + |\mathcal{C}_{t_a}| \log |\mathcal{C}_{t_a}|)$.

Following this, the bi-level optimization process alternates between updating the generator and the surrogate model. The generator's complexity, derived primarily from the attention mechanism and diversity loss calculation, totals $O(B \cdot K^{(pool)} \cdot (d + B))$, where $B$ is the batch size and $K^{(pool)}$ is the candidate pool size. For the surrogate model, each update involves graph convolutions. Given the sparsity of injected edges relative to the total edges $|\mathcal{E}|$, this approximates to $O(L \cdot |\mathcal{E}| \cdot d)$, where $L$ is the number of GNN layers. Assuming $N$ inner updates per outer iteration, the total training complexity per iteration is $O(N \cdot L \cdot |\mathcal{E}| \cdot d + B \cdot K^{(pool)} \cdot (d + B))$, where $d$ denotes the node feature dimension.

\section{Other training parameters}
\label{appendix_sec:other training parameters}
Table \ref{tab:combined_params_simplified} summarizes the training settings, where the trigger size is fixed at 3. The chosen trigger types are Author (ACM), Paper (DBLP), and Director (IMDB). Experiments were executed on an NVIDIA RTX 3090 GPU (24GB), repeated at least three times, and convergence was ensured for all experiments.

\end{document}